\documentclass[pdflatex,sn-nature]{sn-jnl}

\usepackage{graphicx}%
\usepackage{multirow}%
\usepackage{amsmath,amssymb,amsfonts}%
\usepackage{amsthm}%
\usepackage{mathrsfs}%
\usepackage[title]{appendix}%
\usepackage{xcolor}%
\usepackage{textcomp}%
\usepackage{manyfoot}%
\usepackage{booktabs}%
\usepackage{algorithm}%
\usepackage{algorithmicx}%
\usepackage{algpseudocode}%
\usepackage{listings}%

\theoremstyle{thmstyleone}%
\theoremstyle{thmstyletwo}%

\theoremstyle{thmstylethree}%

\begin{document}

\title[MaRaI]{Metadata Supervised Imaging Representations for Modelling and Controlling Acquisition Variability}


\author*[1]{\fnm{Mehmet Yigit} \sur{Avci}}\email{yigit.avci@kcl.ac.uk}

\author[1]{\fnm{Pedro} \sur{Borges}}\email{pedro.borges@kcl.ac.uk}

\author[1]{\fnm{Virginia} \sur{Fernandez}}\email{virginia.fernandez@kcl.ac.uk}

\author[1]{\fnm{Natalia} \sur{Glazman}}\email{natalia.glazman@kcl.ac.uk}

\author[2]{\fnm{Paul} \sur{Wright}}\email{paul.w@ucl.ac.uk}

\author[3]{\fnm{Mehmet} \sur{Yigitsoy}}\email{mehmet.yigitsoy@gmail.com}

\author[1]{\fnm{Sebastien} \sur{Ourselin}}\email{sebastien.ourselin@kcl.ac.uk}

\author[1]{\fnm{M. Jorge} \sur{Cardoso}}\email{m.jorge.cardoso@kcl.ac.uk}

\affil[1]{\orgdiv{School of Biomedical Engineering \& Imaging Sciences}, \orgname{King's College London}, \orgaddress{\city{London}, \country{UK}}}

\affil[2]{\orgname{University College London}, \orgaddress{\city{London}, \country{UK}}}

\affil[3]{\orgname{deepc GMBH}, \orgaddress{\city{Munich}, \country{Germany}}}
\abstract{Biomedical imaging data exhibit substantial acquisition variability, where identical biological structures can appear markedly different due to differences in imaging devices, acquisition protocols, sites, and reconstruction settings. Consequently, learned representations often entangle underlying biological information with acquisition-dependent appearance, limiting interpretability, generalisation, and clinical deployment. We show that these sources of variation can be disentangled by jointly modelling medical images and acquisition metadata. Using large-scale clinical brain MRI data as a case study, we learn representations that disentangle anatomical structure from contrast-dependent appearance. The resulting framework enables the organisation of heterogeneous imaging protocols, sequence understanding, the detection of image–metadata inconsistencies and imaging artifacts, while preserving biologically relevant anatomical features across diverse acquisitions. Building on these disentangled representations, it further supports generative and translational capabilities, performing both metadata-conditioned synthesis of realistic 3D brain MRIs and anatomy-preserving harmonisation for cross-modality and cross-site adaptation. Our findings demonstrate that acquisition variability is a structured component of the imaging process that can be modeled, audited, synthesised, and controlled, establishing a foundation for acquisition-aware representation learning in large-scale biomedical imaging.}




\maketitle

\section{Introduction}\label{sec1}

Biomedical imaging is a cornerstone of modern healthcare and life sciences, enabling non-invasive visualisation and quantitative assessment of biological structure, function, and pathology across diverse clinical and research applications. However, despite rapid progress in machine learning for medical image analysis, robust deployment of automated imaging models across institutions remains challenging. A key limitation is acquisition shift, where differences in imaging devices, acquisition protocols, and reconstruction settings induce substantial changes in image appearance without corresponding changes in the underlying biology \cite{kushol2023mri}. Unlike natural images, biomedical images are not solely determined by the object being imaged; instead, image appearance reflects a complex interaction between biological structure and the acquisition process. As a result, the same images can exhibit markedly different intensity distributions, textures, and contrast under different acquisition conditions, while visually similar scans may originate from distinct protocols rather than shared biology \cite{yan2020mri,Kelly2022,SendraBalcells2022DomainGeneralization}. When learned representations entangle acquisition-specific factors with anatomical content, downstream models become sensitive to domain shifts, limiting their generalisability and hindering reliable deployment across scanners, institutions, and patient populations \cite{posselt2024flairsimulation,glocker2019machine}.

This challenge reflects a broader limitation of current MRI representation learning. Existing approaches address acquisition variability through two main strategies. The first is domain generalisation, which trains models to learn representations that are invariant to scanner or site identity, typically through adversarial objectives that suppress domain-discriminative features~\cite{dinsdale2021, kamnitsas-domain-adaptation}. The second is image harmonisation, which maps images from one acquisition setting into the appearance of another~\cite{YANG2026104066,abbasi2024harmonization,9557808}. Methods span statistical batch-effect correction~\cite{combat}, learned image-to-image translation, including cycle-consistency networks~\cite{zhu2020unpairedimagetoimagetranslationusing}, paired domain translation~\cite{Modanwal2020,dewey2020}, unpaired methods~\cite{WU2026103849}, and disentangled latent space approaches~\cite{ZUO2021118569,haca3,ROCA2025103388,ouyang2021representation}. Both strategies rest on a shared assumption that acquisition-related variation is a nuisance factor to be removed or ignored. However, in MRI, this variability is a structured and physically meaningful consequence of the imaging process itself. By ignoring this underlying physical process, current methods rarely provide representations that explicitly separate anatomical structure from acquisition-dependent appearance, limiting cross-site comparison, interpretation of acquisition effects and systematic quality control.

Furthermore, recent advances in generative modelling have enabled realistic synthesis of brain MRI and generation of common contrasts such as T1-weighted and T2-weighted images \cite{zhao2026maisi,pinaya2022brain,liu2022flow,PENG2024103325}. However, specifying a high-level target modality does not specify the acquisition that produced it: the T1-weighted or T2-weighted appearance can arise from different combinations of acquisition parameters, scanners and reconstruction settings. Consequently, existing generative models therefore largely control the generic type of contrast being generated, rather than the specific acquisition conditions underlying its appearance.

Crucially, much of the information governing acquisition-dependent appearance is already recorded during routine clinical imaging. Every MRI scan is accompanied by DICOM metadata that describes the physical and technical parameters underlying image formation, including echo time, repetition time, sequence type, scanner vendor, and reconstruction settings \cite{DICOMStandard}.  Prior work has leveraged metadata for sequence and body part classification \cite{gauriau2020dicom,liang2021mri,du2020radnet,mcdaniel2022contrast, Jonske2022}, metadata-conditioned reconstruction \cite{contextmri} and scanner harmonisation \cite{tumsyn}, while multimodal contrastive learning has demonstrated the broader utility of aligning medical images with auxiliary information sources \cite{oct-clip,biomedclip,11515314}. However, metadata itself is often incomplete, inconsistent or manually specified in large retrospective archives \cite{metadatagueld}, creating a dual challenge: metadata provides a rich description of acquisition variability, yet its reliability cannot be assumed. Understanding the relationship between image appearance and acquisition metadata is therefore important not only for harmonisation and representation learning, but also for auditing data quality at scale.

\begin{figure*}[t]
    \centering
    \includegraphics[width=\textwidth]{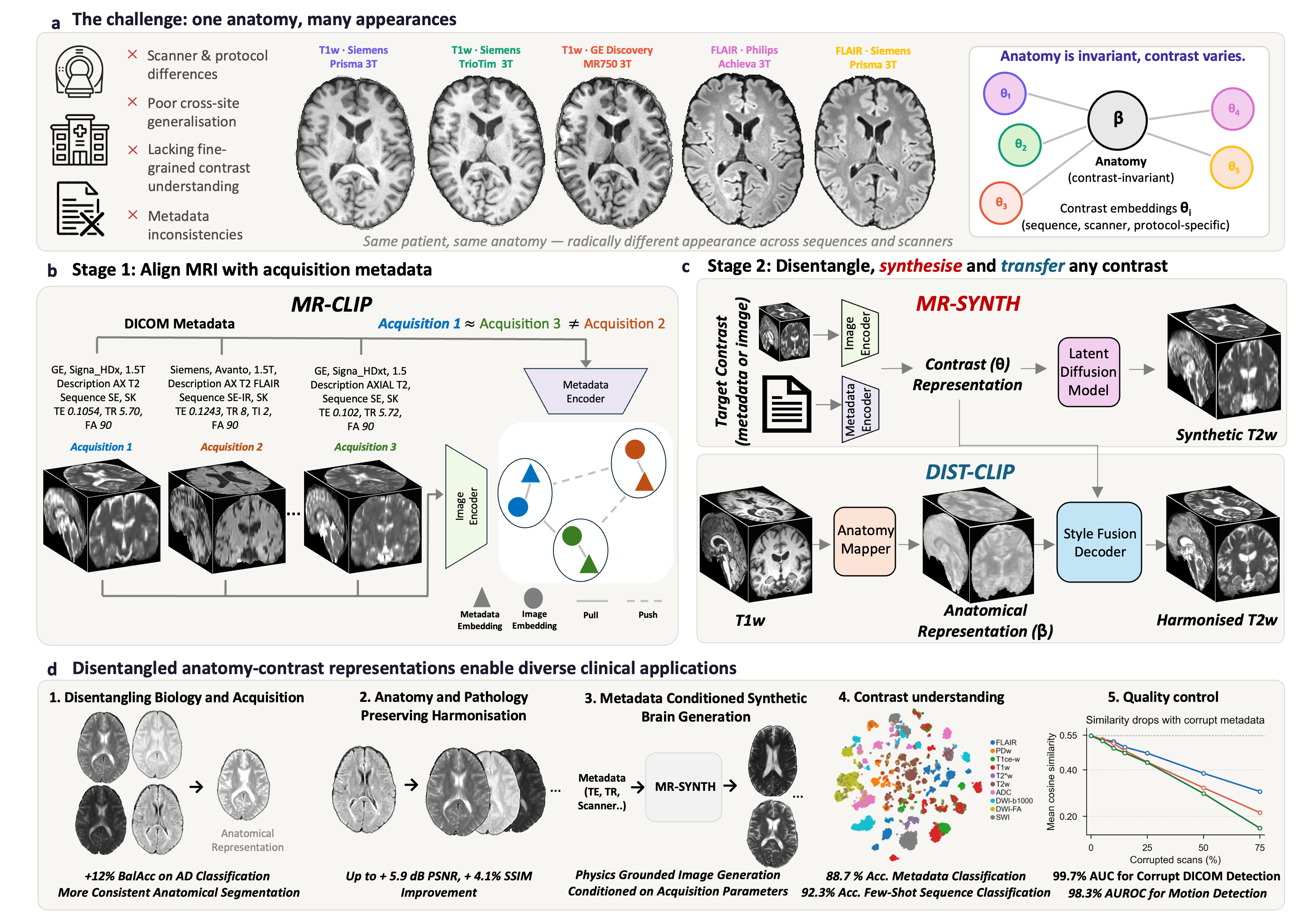}
    \caption{
    \textbf{Disentangled MRI representation learning under acquisition variability.}
    \textbf{a,}
    MRI appearance varies substantially across scanners, protocols and contrasts despite shared underlying anatomy, motivating separation of anatomy from acquisition-dependent appearance.
    \textbf{b,}
    Image--metadata contrastive alignment pairs each MRI volume with a natural-language prompt built from its DICOM tags in a shared embedding space. 
    \textbf{c,}
    A synthesis and harmonisation framework built on pretrained contrast embeddings supports both contrast transfer and brain image synthesis. For harmonisation, each volume is disentangled into an anatomical map ($\beta$) and a contrast code ($\theta$), allowing the anatomy of a source scan to be combined with a target contrast. The same contrast representations can also condition a generative model to synthesise new brains from target acquisition metadata.
    \textbf{d,}
    Factorised anatomical and contrast representations enable diverse downstream applications, including anatomically consistent harmonisation, cross-site alignment, sequence understanding and quality control.
    }
    \label{fig:overview}
\end{figure*}

Here we introduce \textbf{MaRaI} (\textbf{M}ultimodal \textbf{A}cquisition-aware \textbf{R}adiology \textbf{AI}), a framework that treats DICOM metadata not as auxiliary labels but as structured supervision for learning representations in which anatomy and acquisition are separable (Fig.~\ref{fig:overview}). MaRaI has three components that share a common representation space. \textbf{MR-CLIP}~\cite{avci2025mrclipefficientmetadataguidedlearning,avci2025metadataaligned3dmrirepresentations} aligns MRI volumes with natural-language prompts constructed from DICOM acquisition tags, using a supervised contrastive objective~\cite{supcon} over discretised metadata groups to pull together volumes that share the same acquisition protocol regardless of anatomy or subject identity. The resulting embeddings capture contrast as a structured variable, supporting downstream tasks, including cross-modal retrieval, few-shot sequence classification and archive-level quality control to detect both metadata-image inconsistency and imaging artifacts with an external artifact detection head. \textbf{DIST-CLIP}~\cite{avci2025distcliparbitrarymetadataimage} uses these embeddings to factorise each scan into an acquisition-invariant anatomical map $\beta$ and a contrast representation $\theta$, extending and unifying disentangled harmonisation methods~\cite{ouyang2021representation,haca3,dewey2020}. This factorisation enables image harmonisation, in which the anatomical representation of a source scan is combined with a target contrast representation to transfer acquisition-dependent appearance while preserving anatomy. Lastly, \textbf{MR-SYNTH}, a 3D latent diffusion model, uses contrast representations as conditioning to generate new brain MRIs with an acquisition-dependent appearance specified directly through DICOM metadata. Together, they provide complementary routes for controlling acquisition-dependent appearance: cross-modality and cross-site harmonisation of an existing image through disentangled anatomy and contrast, and synthesis of new images directly from acquisition metadata.
 
 Our main contributions are:
\begin{itemize}
\item \textbf{Anatomy–contrast disentanglement and image harmonisation.} We disentangle anatomical structure from acquisition-dependent contrast and quantify the degree of resulting separation across acquisition settings. The resulting anatomical representations reduce acquisition-driven variability in tissue morphometry across protocols and improve cross-vendor Alzheimer’s disease classification. Building on this factorisation, a novel style transfer architecture injects contrast information into the anatomical representations, enabling precise, layer-wise control of acquisition-dependent appearance without altering the underlying anatomy. Image-guided and metadata-conditioned modes share a single set of trained weights, yielding a unified model that performs high-fidelity harmonisation across structural brain MRI contrasts. 

\item \textbf{Metadata-conditioned brain MRI synthesis.} To our knowledge, we introduce the first 3D latent diffusion model for brain MRI synthesis that controls acquisition-dependent contrast directly from MRI acquisition parameters and scanner specifications through disentangled contrast representations derived from DICOM metadata, without explicit MRI physics simulations. The resulting images exhibit appearance consistent with expected MRI physics and capture scanner-specific variation in the learned embedding space.

\item \textbf{Metadata-driven contrast-aware representation learning.} We introduce a metadata discretisation strategy that converts heterogeneous DICOM acquisition parameters into structured supervision signals, enabling large-scale supervised contrastive learning. The resulting MR-CLIP embeddings support retrieval, few-shot classification, and quality control, outperforming supervised baselines and demonstrating that acquisition metadata can effectively structure large-scale clinical archives.
\end{itemize}


Together, we unify these capabilities within the \textbf{MaRaI} framework, using routine acquisition metadata to learn representations that separate anatomy from acquisition-dependent contrast, harmonise existing images across acquisition settings, generate images with controllable acquisition-dependent appearance, and screen large clinical imaging archives for metadata–image inconsistencies and artifacts. 

\section{Results}\label{sec2}

\subsection{Anatomical representations suppress acquisition-dependent variability}\label{sec:anatomy}

DIST-CLIP decomposes each scan into two complementary  representations: an anatomical map $\beta$ that captures  tissue structure independently of how that structure was 
imaged, and a contrast code $\theta$ that encodes  acquisition-dependent appearance. For this factorisation to  be meaningful, $\beta$ must preserve structural information like sulcal geometry, ventricular shape, tissue boundaries  that is consistent across protocols while relegating  sequence-specific intensity characteristics to $\theta$.  We evaluated this claim through two experiments: assessment of morphometric  stability across registered multi-contrast acquisitions,  and cross-vendor Alzheimer's disease (AD) classification on ADNI~\cite{jack2008adni}.

\begin{figure*}[h]
    \centering
    \includegraphics[width=\textwidth]{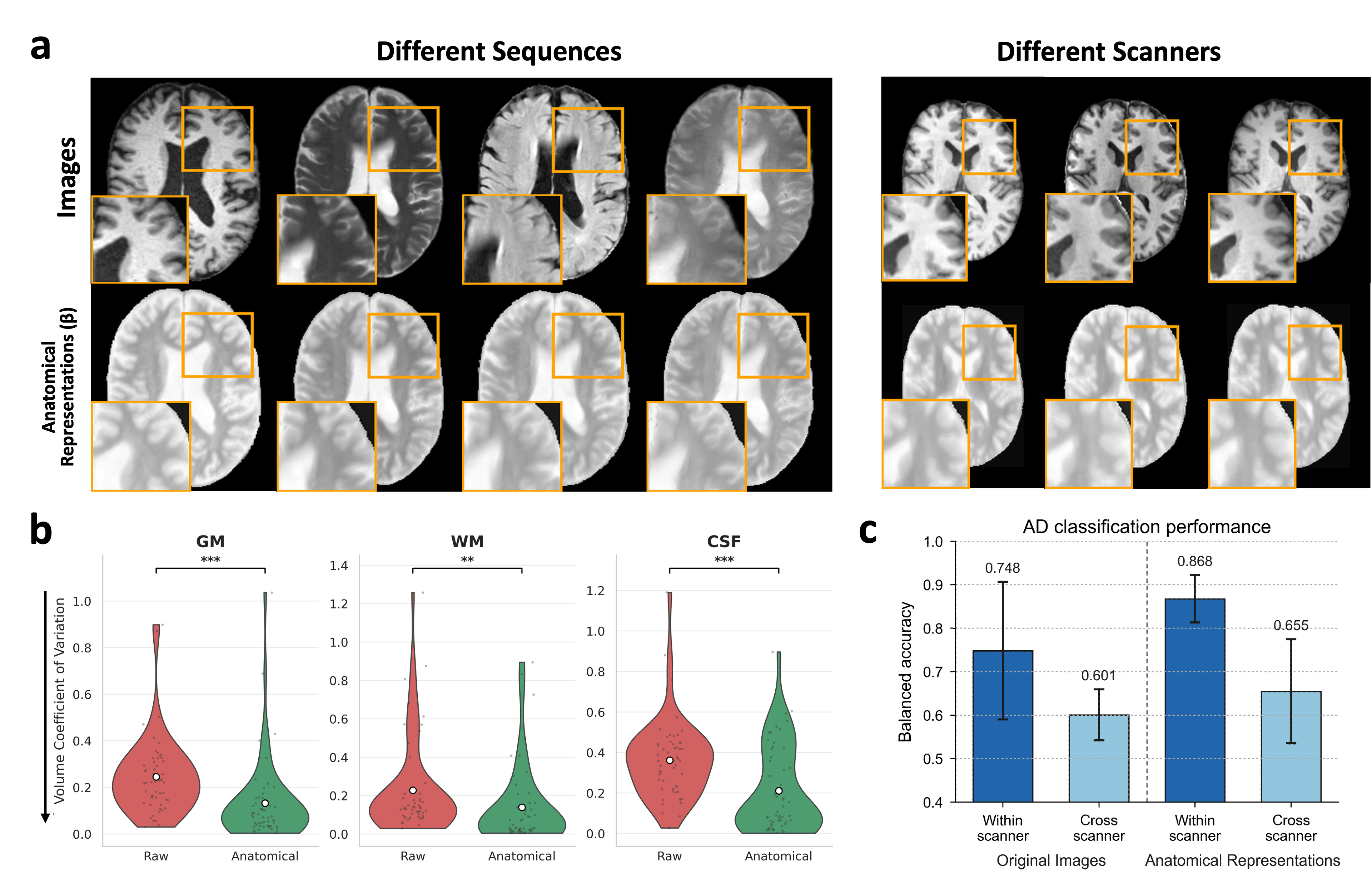}
    \caption{
    \textbf{Anatomical representations suppress acquisition-dependent variability.}
    \textbf{a,}
    Axial slices from raw MRI acquisitions from different sequences and scanners exhibit substantial appearance variability despite shared underlying anatomy. The proposed framework maps heterogeneous acquisitions into anatomically consistent representations while preserving fine-grained structural boundaries.
    \textbf{b,}
    Coefficient of variation analysis of tissue-specific segmentation volumes, evaluated across different sequences with at least three acquisitions, demonstrates reduced variability in anatomical representations compared with raw MRI images, indicating suppression of acquisition-dependent variation while preserving shared anatomical structure.
    \textbf{c,}
    On ADNI, Alzheimer's disease classifiers trained on one scanner generalise better to data from another scanner with anatomical representations than with raw MRI, indicating reduced acquisition variability without loss of diagnostic signal.
    }
    \label{fig:anatomy}
\end{figure*}

Qualitatively, raw slices from the same subject varied substantially in contrast across sequences and scanners, whereas our anatomical representations showed aligned sulcal and ventricular anatomy (Fig.~\ref{fig:anatomy}a). Importantly, we do not expect anatomical representations from different sequences to become identical, as each acquisition retains sequence-specific signal. Instead, the objective is to suppress non-biological acquisition variability while preserving meaningful anatomical content, reflected as patchwise structural consistency rather than exact intensity matching. The degree of disentanglement is quantified in Supp. Fig.~\ref{fig:beta_disentanglement}, showing that $\beta$ captures predominantly anatomical information, whereas $\theta$ captures predominantly acquisition-dependent contrast information. Furthermore, we applied SPM12 unified segmentation~\cite{ashburner2005unified} to raw images and matched $\beta$-maps from sessions with $\geq3$ acquisitions (Fig.~\ref{fig:anatomy}b). Mean Dice overlap across GM, WM and CSF was similar for raw and anatomical images (0.608 versus 0.582; all tissue-wise comparisons not significant), whereas mean inter-sequence volume CV was significantly lower for anatomical maps (0.161 versus 0.279; $p<0.05$ for GM, WM and CSF). Thus, anatomical outputs remained tissue-consistent while reducing protocol-driven variability.

We next evaluated whether this stability improves downstream prediction under scanner shift. On ADNI, AD classifiers trained on Siemens scans and tested without fine-tuning on GE scans achieved higher balanced accuracy with $\beta$-inputs than raw MRI (0.65 vs.\ 0.60, $p>0.05$). Within-scanner evaluation also showed consistent improvement with $\beta$-inputs, with an approximately 12\% gain ($p<0.05$) (Fig.~\ref{fig:anatomy}c). Overall, the anatomy pathway improves cross-vendor robustness while preserving clinically relevant signal.

\subsection{Anatomy-preserved unified harmonisation across contrasts and scanners}\label{sec:harmonisation}

\begin{figure}[!ht]
    \centering
    \includegraphics[width=0.8\textwidth]{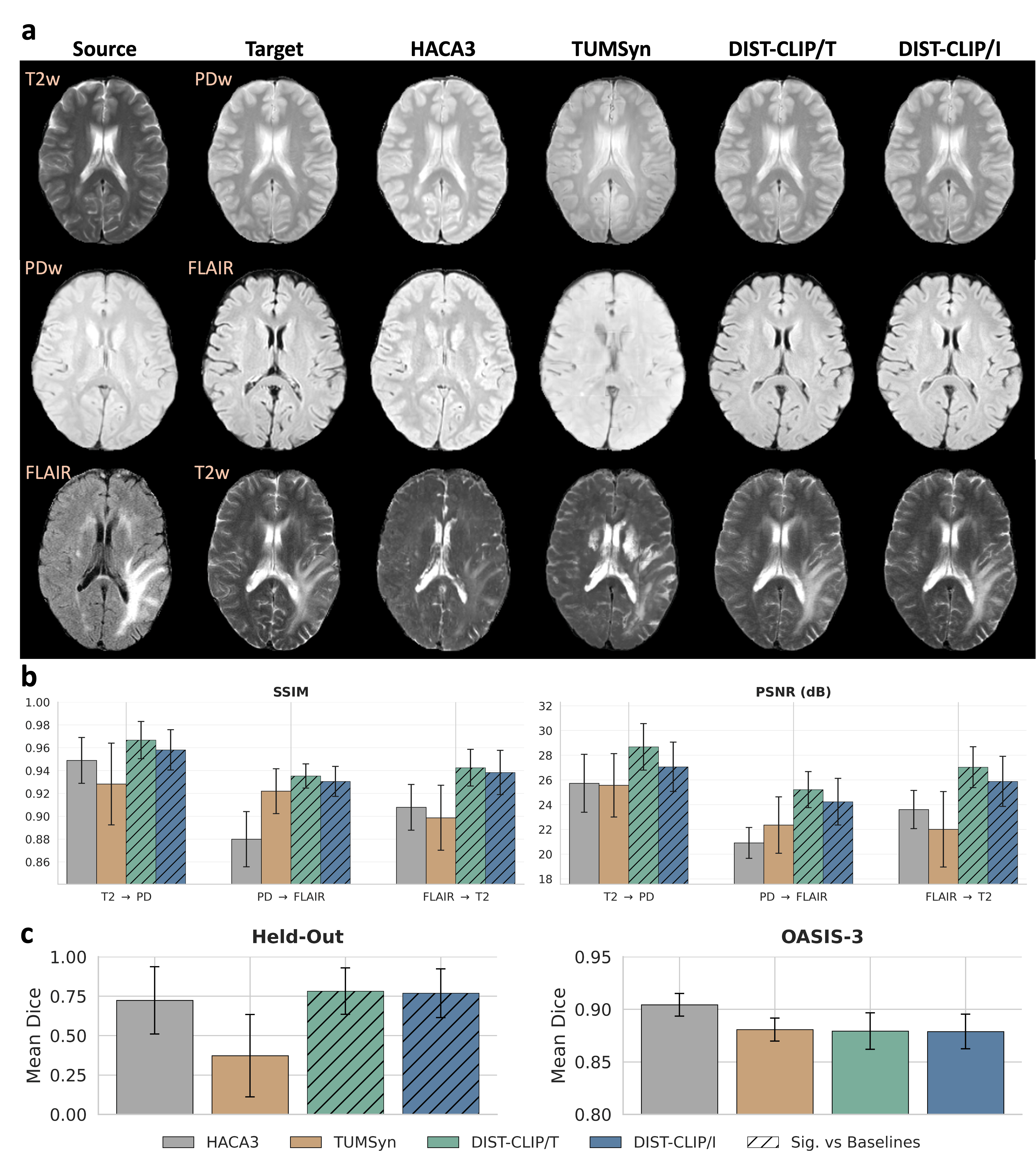}
    \caption{
        \textbf{Cross-contrast harmonisation and segmentation consistency.}
        \textbf{a,}
        Qualitative T2w$\rightarrow$PDw, PDw$\rightarrow$FLAIR and FLAIR$\rightarrow$T2w examples (source, target, HACA3, TUMSyn, DIST-CLIP/T, DIST-CLIP/I).
        \textbf{b,}
        Mean SSIM and PSNR ($\pm$s.d.) on the same three translation directions.
        \textbf{c,}
        Mean Dice ($\pm$s.d.) between SynthSeg~\cite{synthseg} parcellations of reference T2w volumes and T1w$\rightarrow$T2w harmonised outputs on held-out clinical test data ($n=20$ subjects) and OASIS-3 ($n=14$ subjects). Hatched bars denote methods that achieve statistically significant improvements over both HACA3 and TUMSyn ($p<0.05$).
    }
    \label{fig:cross_contrast}
\end{figure}

Harmonisation directly follows from anatomy-contrast disentanglement: once anatomical structure is separated from acquisition-dependent contrast, new image appearances can be generated by manipulating the contrast component. In this setting, a source scan provides the anatomical content, while the target contrast is specified either through acquisition metadata or a reference image, enabling synthesis of cross-contrast outputs. We trained on registered multi-contrast pairs and compared against image-based HACA3~\cite{haca3} and metadata-based TUMSyn~\cite{tumsyn} under matched preprocessing and evaluation. Pixel-level fidelity was quantified with SSIM and PSNR on held-out source$\rightarrow$target pairs (758 bidirectional pairs, across the four main contrasts; Supplementary Table~\ref{tab:harm_test_pairs}; all directions in Supplementary Fig.~\ref{fig:supp_metrics} and ~\ref{fig:supp_metrics_masked}). For the representative directions in Fig.~\ref{fig:cross_contrast}a,b, DIST-CLIP/I and DIST-CLIP/T achieved the strongest SSIM/PSNR, reaching 30.3\,dB and 0.969 SSIM on PDw$\rightarrow$T2w. Qualitatively, HACA3 produced locally sharp results but occasionally generated implausible tissue intensities, particularly in FLAIR where ventricular signal was not faithfully preserved. TUMSyn reduced high-frequency noise but often caused oversmoothing and grid-like artefacts, with inconsistent target-domain appearance across volumes.

For volumetric consistency, we trained a DIST-CLIP-2.5D variant without the Anatomy Mapper due to memory constraints. A single 3D MR-CLIP contrast embedding was shared across slices, with a shared 2D decoder performing slice-wise harmonisation. This improved sagittal and coronal coherence, at the cost of some cortical oversmoothing (Supplementary Fig.~\ref{fig:supp_visual}).

We next asked whether improved pixel fidelity translates into better preservation of anatomically meaningful structure. Using SynthSeg~\cite{synthseg}, we compared segmentations of T1w$\rightarrow$T2w harmonised outputs with corresponding reference T2w scans (mean Dice across structures; Fig.~\ref{fig:cross_contrast}c; qualitative overlays in Fig.~\ref{fig:qualitative_synthseg_overlays}). On held-out clinical data ($n=20$ subjects), DIST-CLIP achieved the highest Dice (0.78), significantly outperforming HACA3 (0.70; $p<0.01$) and TUMSyn (0.38; $p<0.001$). On OASIS-3 ($n=14$ subjects), a more homogeneous research dataset on which HACA3 and TUMSyn were trained but DIST-CLIP was not, all methods achieved similarly high Dice (0.88--0.90).

On the external ON-Harmony travelling-heads dataset, DIST-CLIP/I and DIST-CLIP/T achieved the highest reconstruction quality (SSIM and PSNR, $p<0.001$ vs.\ all baselines) while substantially reducing site identifiability without degrading brain-age prediction. DIST-CLIP/I reduced site classification accuracy from 0.79 to 0.42 while maintaining brain-age MAE (8.89 vs.\ 8.80 years for unharmonised data), outperforming the baselines in jointly removing acquisition-specific variation and preserving age-related anatomical information (Supp. Fig. \ref{fig:cross_scanner}).

Finally, zero-shot evaluation on OASIS-3, a dataset not used during the training of DIST-CLIP whereas the baselines were trained on it, showed that DIST-CLIP produces anatomically plausible outputs with highly competitive SSIM and PSNR (Supplementary Fig. \ref{fig:supp_oasis}).

For harmonisation to be meaningful in clinical imaging, it must not alter pathological structures that are relevant for downstream analysis. We therefore evaluate pathology preservation using independent, unseen cohorts containing white matter hyperintensities (WMHs) and brain tumours, providing quantitative and qualitative assessments of focal pathology under harmonisation.

\begin{figure}[ht]
\centering
\includegraphics[width=0.99\linewidth]{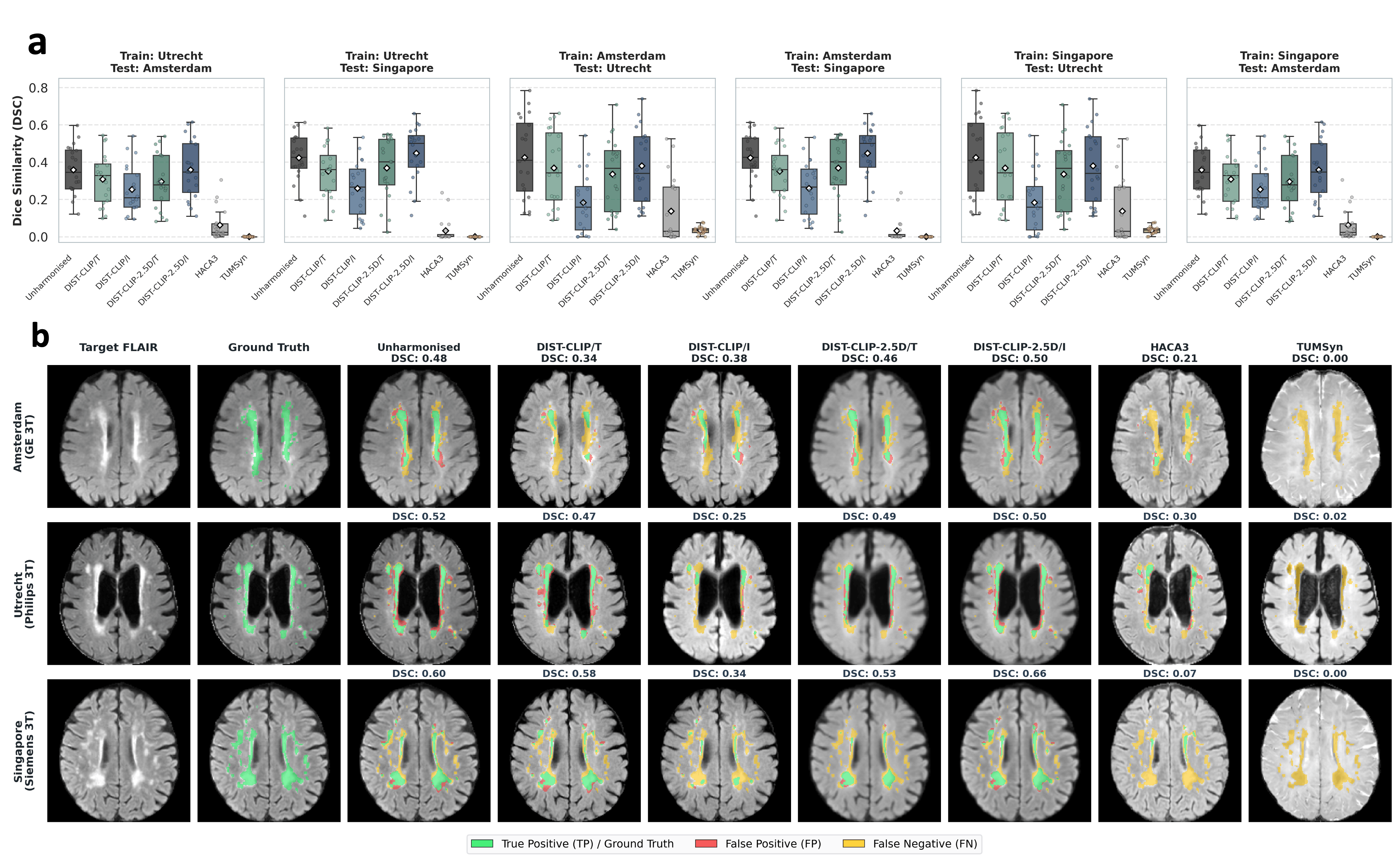}
\caption{    \textbf{Cross-site WMH segmentation.}
\textbf{a, } An nnU-Net segmenter trained exclusively on native scans from the source site was evaluated on raw unharmonised target scans and harmonised outputs. Boxplots show the median and interquartile range, with white diamonds indicating the mean Dice score. \textbf{b, } Visual harmonised examples with ground truth segmentations.}
\label{fig:wmh}
\end{figure}

The WMH Challenge \cite{Kuijf2019} provides a particularly challenging test, with substantial differences between sites. We trained an nnU-Net \cite{isensee2021nnu} segmentation model on native scans from one centre and evaluated it on scans from each of the other centres, covering all six directed site pairs. For each target site, we compared segmentation performance on the raw, unharmonised scans with the corresponding harmonised outputs from DIST-CLIP, HACA3 and TUMSyn. As shown in Fig.~\ref{fig:wmh}a, DIST-CLIP-2.5D generally maintains or improves WMH segmentation performance relative to the raw target scans across unseen acquisition domains, indicating that harmonisation does not compromise the reconstruction of focal pathology. In contrast, HACA3 and TUMSyn show substantial performance degradation in several cross-site settings, primarily reflecting difficulties in translating FLAIR appearance across unseen acquisition domains as can be seen in qualitative examples in Fig. ~\ref{fig:wmh}b. Further qualitative examples of brain tumour harmonisation using BraTS dataset \cite{baid2021rsna} and additional held-out pathological cases are provided in Supplementary Figs. \ref{fig:supp_brats} and \ref{fig:pat}.

\subsection{Metadata-conditioned synthesis of high-fidelity 3D brain MRI}

We next asked whether acquisition metadata could control the appearance of newly generated brain MRI. We therefore developed MR-SYNTH, a 3D latent diffusion model conditioned on MR-CLIP representations of acquisition metadata, to generate whole-brain MR volumes aligned with specified sequence and scanner conditions.

\begin{figure}[h]
\centering
\includegraphics[width=\linewidth]{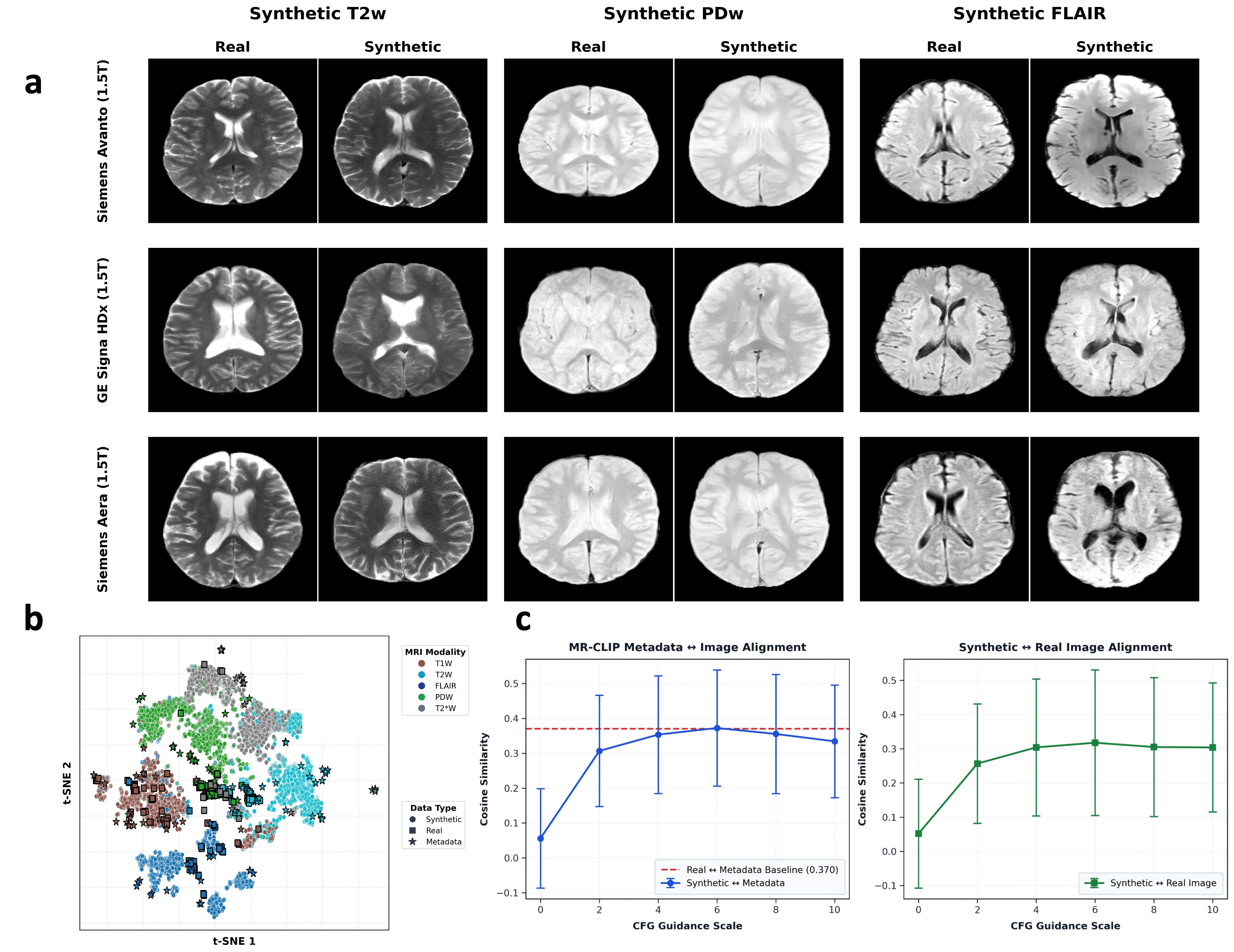}
\caption{\textbf{Metadata-conditioned 3D brain MRI synthesis and conditioning fidelity.}
\textbf{a}, Qualitative comparison of real clinical acquisitions and synthetic volumes with same metadata prompts across T2w, PDw and FLAIR acquisitions on Siemens Avanto 1.5T, GE Signa HDx 1.5T and Siemens Aera 1.5T scanners. Synthetic volumes reproduce sequence-specific contrast, including hyperintense CSF in T2w, subtle tissue contrast in PDw and CSF suppression in FLAIR, while maintaining plausible anatomy.
\textbf{b}, t-SNE projection of MR-CLIP embeddings for real acquisitions (squares), synthetic volumes (circles) and conditioning metadata (stars) across five modalities. Synthetic volumes co-cluster with both real acquisitions and their corresponding metadata.
\textbf{c}, Effect of classifier-free guidance (CFG) on conditioning fidelity. Image--metadata cosine similarity approaches the real-image baseline ($0.370 \pm 0.17$) at CFG $4$--$6$. Synthetic-to-real image similarity is highest at CFG 6 ($0.318 \pm 0.21$). Error bars denote s.d. across $N=30$ test subjects.}
\label{fig:main_results}
\end{figure}

Across sequence types and scanners, MR-SYNTH generated anatomically plausible brain volumes with acquisition-dependent appearance consistent with the conditioning metadata (Fig.~\ref{fig:main_results}a). T2w, PDw and FLAIR generations showed the expected differences in tissue and CSF signal, including CSF hyperintensity in T2w, more subtle tissue contrast in PDw and CSF suppression in FLAIR. These characteristics were reproduced across different scanners, indicating that the model captured acquisition-dependent variation beyond broad sequence type.

We next assessed whether the generated images remained aligned with their conditioning metadata and with real clinical acquisitions in the MR-CLIP embedding space. Across five modalities, synthetic volumes co-clustered with real acquisitions and their corresponding metadata (Fig.~\ref{fig:main_results}b), indicating that the acquisition information specified by the conditioning signal was reflected in the generated images. Image--metadata similarity increased with Classifier Free Guidance (CFG) and reached the real-image baseline ($0.370 \pm 0.17$) at CFG value of 6 (Fig.~\ref{fig:main_results}c), while synthetic-to-real image similarity was highest at CFG $6$ ($0.318 \pm 0.21$). Also, the model exhibited behaviour consistent with expected MRI physics, with image appearance changing systematically as acquisition parameters were continuously varied (Supp. Fig.~\ref{fig:physics_te_tr_sweeps}). Synthetic images remained clustered by modality while separating scanner-specific variations within each modality, consistent with the structure observed in real acquisitions (Supp. Fig. \ref{fig:scanner_tsne}), qualitative examples are in Supp. Fig. \ref{fig:real_vs_synth_matrix}.

\subsection{Metadata-guided representations organise acquisition-dependent latent structure}\label{sec:metadata}

MR-CLIP is designed to understand and encode image contrast by aligning MRI scans with acquisition metadata derived from DICOM tags. The motivation is that MRI appearance is fundamentally determined by acquisition parameters, yet these factors are often treated as noisy labels or ignored in representation learning. We address this by discretising numerical metadata and grouping acquisitions into contrast definitions as described in Section~\ref{contrast-group}. Using these discretised acquisition metadata as supervision signals, MR-CLIP learns embeddings in which image similarity is structured by imaging protocol rather than manually defined broad sequence labels.

\begin{figure}[!ht]
    \centering
    \includegraphics[width=\linewidth]{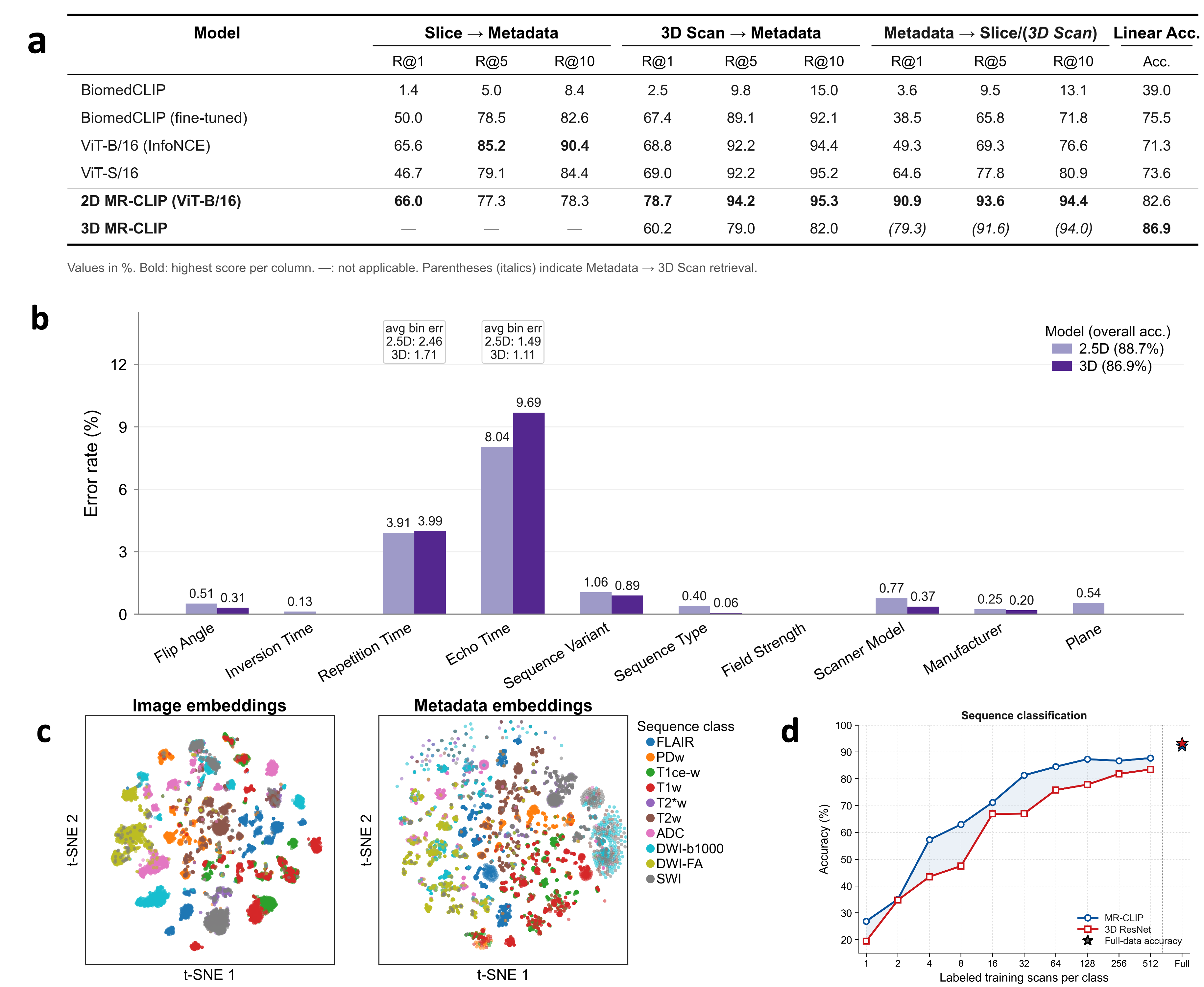}
    \caption{
    \textbf{Metadata-guided representations organise acquisition-dependent latent structure}
    \textbf{a,}
    Cross-modal retrieval (recall at ranks 1, 5 and 10) for slice-to-metadata, 3D scan-to-metadata and metadata-to-slice matching, with linear-probe accuracy for metadata classification. Highest value per column is shown in bold.
    \textbf{b,}
    Per-field linear-probe error rates (\%) for 2.5D and 3D MR-CLIP across ten DICOM tags; inset values report mean bin-level error for discretised echo time (TE) and repetition time (TR).
    \textbf{c,}
    t-SNE of image embeddings (left) and metadata embeddings (right) coloured by sequence class.
    \textbf{d,}
    Few-shot sequence classification accuracy versus labelled training scans per class, comparing a linear classifier on frozen MR-CLIP image embeddings with a supervised 3D ResNet; star denotes full-data accuracy (MR-CLIP significantly outperforms ResNet across few-shot levels, $p < 0.0001$).
    }
    \label{fig:metadata}
\end{figure}

Cross-modal retrieval on the held-out test set (Fig.~\ref{fig:metadata}a) quantifies whether embeddings organise acquisitions by protocol. Two-dimensional MR-CLIP reached 66.0\% recall@1 for slice-to-metadata matching and 90.9\% for metadata-to-slice matching. At volume level, scan-to-metadata recall@1 reached 78.7\% when slice embeddings were aggregated and 60.2\% with the native three-dimensional encoder. MR-CLIP consistently outperformed frozen and fine-tuned BiomedCLIP, and in most configurations achieved higher performance than the InfoNCE ViT-B/16 baseline. Metadata-to-slice retrieval exceeded slice-to-metadata retrieval, as expected since many slices share one acquisition description. Incorporating demographic attributes (age and sex) into the 2D model did not yield additional performance gains. Main results used skull-stripped volumes; retaining the skull slightly increased retrieval scores (Supplementary Table~\ref{tab:skull_ablation_retrieval}).

We then trained linear probes to predict all metadata attributes from frozen embeddings (Fig.~\ref{fig:metadata}b). Categorical attributes such as imaging plane were recovered with near-zero error, while 3D MR-CLIP showed the largest errors for TE and TR (1.11 and 1.71 bins; $\sim$11.1 ms and 855 ms), likely reflecting discretisation ambiguity rather than missing contrast information. Overall, 2.5D slice aggregation slightly outperformed 3D, owing to max-vote aggregation across slices.

t-SNE projections of image and metadata embeddings formed well-aligned clusters that emerged naturally according to major MRI sequence families (e.g., T1w, T2w, FLAIR, and diffusion-weighted imaging; Fig.~\ref{fig:metadata}c), with consistent cross-modal clustering extending across scanner manufacturers, scanner models, and acquisition planes (Supplementary Fig.~\ref{fig:tsne_multimodal}). In few-shot sequence classification, a linear classifier on frozen MR-CLIP embeddings significantly outperformed an end-to-end 3D ResNet when labels were scarce ($p < 0.0001$ at all evaluated few-shot levels except 2 shots) and matched it with abundant labels (Fig.~\ref{fig:metadata}d).

\subsection{Automated quality control for metadata inconsistencies and imaging artifacts}\label{sec:qc}

\begin{figure}[ht]
    \centering
    \includegraphics[width=0.9\linewidth]{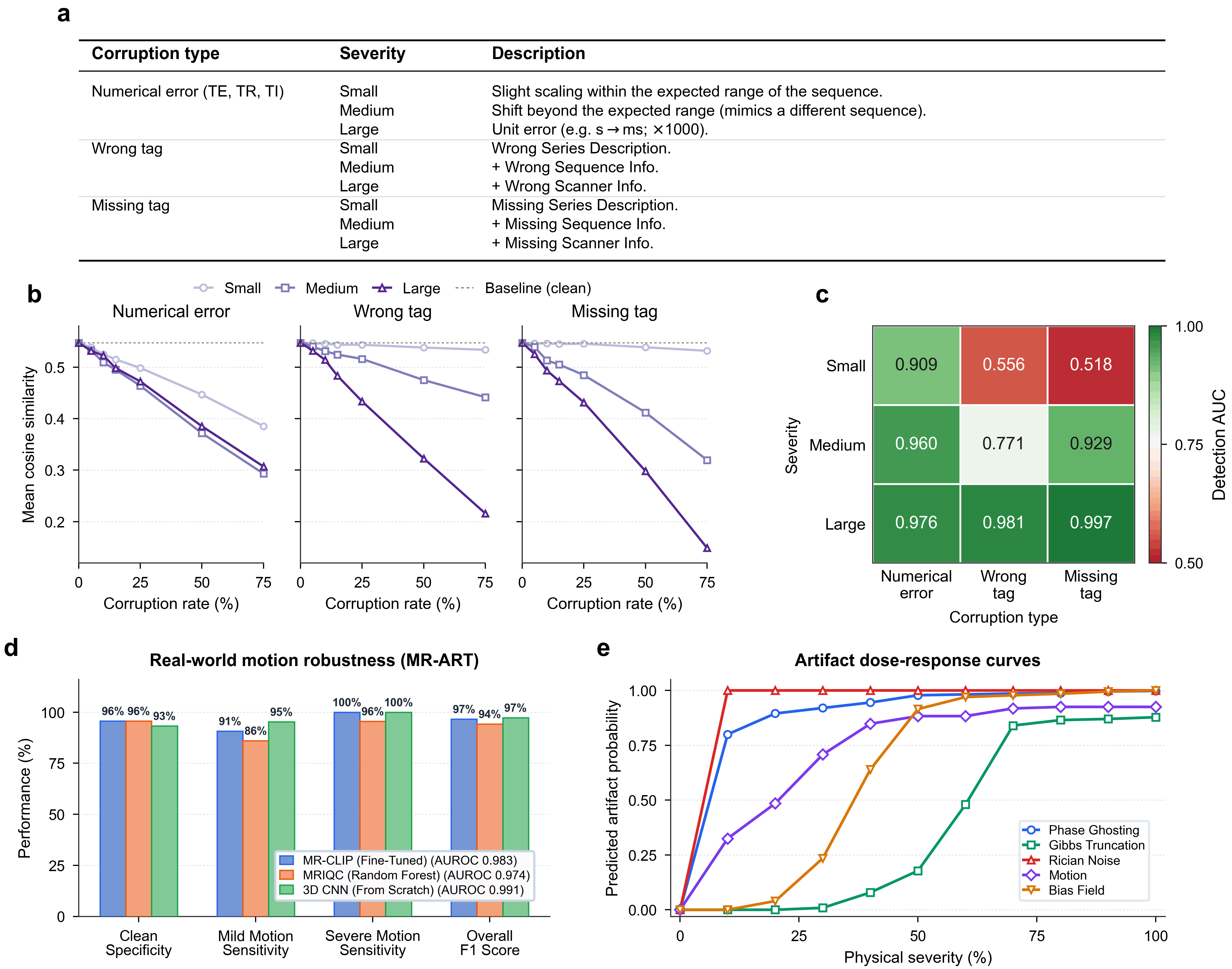}
    \caption{
    \textbf{Detection of DICOM metadata corruption and imaging artifacts with a single model.}
    \textbf{a,}
    Synthetic corruption protocol applied to held-out test metadata, comprising numerical errors; incorrect categorical tags; and missing fields, each at small, medium and large severity.
    \textbf{b,}
    Mean cosine similarity between MR-CLIP image and metadata embeddings as a function of corruption rate (0--75\%) for each corruption type and severity; dotted line denotes clean baseline.
    \textbf{c,}
    Detection AUC at 50\% corruption rate using embedding distance as an anomaly score, evaluated separately for each corruption type and severity (colour scale: 0.5--1.0).
    \textbf{d, }
    Real-world artifact-detection performance on the MR-ART dataset, comparing MR-CLIP (fine-tuned) against MRIQC, and a 3D CNN trained from scratch, reported as clean-scan specificity, sensitivity to mild and severe motion, and overall F1 score.
\textbf{e,}
Dose-response curves showing the model's predicted artifact probability as a function of physical corruption severity (0--100\%) for five TorchIO-simulated artifact types (phase ghosting, Gibbs truncation, Rician noise, motion, bias field).
    }
    \label{fig:qc}
\end{figure}

Clinical imaging archives rely on accurate DICOM metadata and consistent image quality for organisation, retrieval and quality control~\cite{sinha2024mrqa,mrqy,keaveney2025protocolqc}. However, acquisition fields are frequently missing, inconsistent or incorrectly recorded. Because MR-CLIP aligns each volume with a metadata prompt, incorrect metadata should reduce image--metadata similarity even when the image itself appears normal. We therefore corrupted metadata in held-out test volumes at rates from 0\% to 75\%, perturbing numerical parameters, swapping categorical tags, or deleting fields at graded severity (Fig.~\ref{fig:qc}a).

Mean image--metadata cosine similarity decreased monotonically with corruption rate across all corruption classes (Fig.~\ref{fig:qc}b). Numerical timing errors produced the largest decline, consistent with the role of TE, TR and TI in determining image contrast. Wrong categorical tags also showed clear separation by severity, while missing sequence or scanner identifiers had the largest effect at high severity. In contrast, changes to Series Description had little effect because this field is inherently noisy and was excluded from contrast groups during training.

At 50\% corrupted volumes, embedding distance distinguished clean from corrupted metadata with high AUC for severe edits (Fig.~\ref{fig:qc}c): 0.997 for large missing-tag errors, 0.981 for large wrong-tag errors and 0.976 for large numerical unit errors. Subtle missing- and wrong-tag edits were more difficult to detect, whereas small numerical shifts remained reliably detectable (AUC 0.909--0.976). Across corruption rates of 5--75\%, detection AUC increased with both corruption rate and severity, with large edits remaining near-perfect even at low prevalence (Supplementary Fig.~\ref{fig:supp_qc}).

We next evaluated image-quality control using an external artifact-detection head trained jointly with MR-CLIP (Fig. \ref{fig:qc}d,e). On the MR-ART benchmark~\cite{Narai2022}, comprising 148 volumes acquired under controlled movement protocols, the model achieved an AUROC of 0.983 and F$_1$-score of 97\%, with 96\% specificity on clean scans and higher sensitivity to mild and severe motion than MRIQC. Across controlled artefact severity sweeps, the model produced generally graded, increasing responses to motion, ghosting, Gibbs ringing, Rician noise and $B_1$ bias fields, supporting continuous rather than binary quality assessment.


\section{Discussion}\label{sec:discussion}

Multi-centre MRI will remain heterogeneous as scanners are replaced, sequences are shortened for throughput and new acquisition protocols are introduced. MaRaI rests on a simple premise: much of that heterogeneity is not opaque, it is recorded in DICOM tags and coupled to acquisition physics. Learning from metadata at hospital scale, then factorising anatomy from acquisition-dependent appearance offers a practical route to reduce, control and audit acquisition shift.

Anatomical maps stabilise morphometry across protocols and improve cross-vendor classification, indicating that a shared structural representation can be extracted. Harmonisation extends the same factorisation to synthesis: fixing anatomical maps while changing contrast or scanner signature is a constructive proof that tissue boundaries are preserved, and downstream evaluations confirm that harmonised outputs lose scanner identifiability while retaining biological signal. The combination of cross-contrast translation, cross-scanner generalisation, and metadata-only target specification with performance close to image-guided synthesis were not previously combined in one harmonisation framework tied to archive-scale contrast learning. 

MR-SYNTH provides a complementary approach by enabling new brain MRIs to be generated under specified acquisition conditions. This provides a way to simulate images with different sequence and scanner characteristics without requiring additional physical acquisitions or explicit MRI physics simulations. Such controllable synthesis could also augment datasets with underrepresented acquisition settings, enabling systematic studies of acquisition-related domain shifts and their effects on downstream models.

Beyond harmonisation and synthesis, the same contrast embeddings provide a scalable approach for organising and auditing large clinical imaging archives. Retrieval by acquisition similarity enables cohort stratification without manual sequence labelling, while few-shot classification can identify sequences when series descriptions are inconsistent or absent. Metadata--image consistency checks can flag volumes whose image appearance is inconsistent with their recorded DICOM metadata, while a jointly trained artifact-detection head can identify poor-quality images. Such discrepancies and artifacts may compromise cohort definition, downstream analyses, or clinical data curation. This complements existing rule-based quality-control tools~\cite{sinha2024mrqa,mrqy,keaveney2025protocolqc} by prioritising suspicious volumes for targeted human review rather than requiring exhaustive manual inspection. In multi-site studies, these capabilities could support a common acquisition specification by providing both a representation for harmonisation and an independent check that acquired images remain consistent with their recorded acquisition parameters.

The same disentangled representations may also help distinguish biological from acquisition-related sources of demographic signal in MRI. Recent work using DIST-CLIP found that demographic predictability was primarily preserved in anatomy-focused representations, whereas contrast embeddings captured weaker, dataset-specific signal~\cite{avci2026understandingsourcesdemographicpredictability}. Such factorised representations may therefore support both acquisition control and auditing of its contribution to downstream models.

Clinical images reflect more than anatomy and acquisition; pathology remains only partially disentangled in MaRaI. Consequently, harmonisation risks mistaking disease-related changes for acquisition style, while downstream models may confound pathology with residual acquisition shifts. Extending the framework to a tri-factor decomposition, anatomy, acquisition, and pathology, mirrors radiological interpretation more faithfully and represents a natural next step.

Our evidence is strongest for adult brain MRI in routine practice. Extending these principles to pediatric cohorts, non-brain anatomies, quantitative MRI, or other modalities requires further validation across wider clinical demographics and vendors. Methodologically, the harmonisation model operates slice-wise, leaving volumetric consistency unconstrained, and relies heavily on informative DICOM metadata, which can be vulnerable to subtle corruptions or incomplete protocol records. Future priorities include prospective deployment trials, expert reader-based QC, and full 3D volumetric extensions to establish broader translational robustness.

Despite these limitations, the central finding remains: acquisition variability need not be treated solely as an unavoidable source of domain shift. By learning from metadata at hospital scale and separating anatomy from acquisition-dependent appearance, variability can instead become a structured component of the imaging process that can be modelled, audited and controlled. This perspective opens the door to acquisition-aware foundation models that explicitly account for and control acquisition variability, an important capability for deployment in heterogeneous real-world imaging systems and for more robust, equitable, and auditable medical imaging AI.

\section{Methods}\label{sec:methods}

MaRaI was trained and evaluated on de-identified clinical brain MRI data, with external benchmarks on ON-Harmony, ADNI and OASIS-3, WMH Challenge, and BraTS 2021. The pipeline comprises shared preprocessing, metadata grouping and prompt construction, MR-CLIP contrastive pretraining, DIST-CLIP harmonisation and MR-SYNTH synthesis with frozen MR-CLIP encoders, and evaluation protocols for retrieval, harmonisation metrics, downstream readouts, synthesis quality, and metadata quality control. 
\subsection{Datasets and data splits}

\textbf{Data Collection.}
Brain MRI data were retrospectively obtained from the institutional PACS at King’s College Hospital (KCH) and Guy’s and St Thomas’ NHS Foundation Trust (GSTT), comprising 169,634 three-dimensional volumes from 40,005 subjects acquired with diverse scanner and acquisition settings between 2020 and 2022 (Supplementary Fig.~\ref{fig:supp_data}). Data were retrieved using a combination of manual transfer and the DICOM Query-Retrieve plugin for XNAT 1.8. DICOM data were anonymised within XNAT by removing 151 attributes potentially containing identifying information.

\noindent \textbf{CLIP training cohort (MR-CLIP).}
DICOM metadata yielded 21,660 unique acquisition configurations, which were consolidated into 1,415 contrast-aware groups after metadata discretisation (see Section \ref{contrast-group}). Data were split at the subject level into training (60\%), validation (10\%), and test (30\%) sets, with an enlarged test set to facilitate a more robust evaluation of harmonisation on unseen subjects.

\noindent\textbf{Harmonisation cohort (DIST-CLIP).}
For cross-contrast synthesis, we assembled paired brain MRI data by selecting imaging sessions in which a subject underwent at least two distinct MRI contrast acquisitions. The resulting dataset comprised 21,115 volumes from 8,466 subjects across 9,280 imaging sessions. Of these sessions, 91.0\% contained two acquisitions, 7.8\% contained three, and 1.2\% contained four. Models were trained on pairs from the MR-CLIP training partition; quantitative harmonisation results in the main figures and Supplementary Fig.~\ref{fig:supp_metrics} are reported on a fixed held-out test set of source--target pairs from MR-CLIP test set partition (counts in Supplementary Table~\ref{tab:harm_test_pairs}). 

\paragraph{External evaluation cohorts}

\noindent\textbf{ON-Harmony:} A travelling-heads harmonisation study~\cite{warrington2025onharmony} designed to assess scanner- and site-related variability. Healthy participants were scanned across multiple 3T clinical sites using GE, Philips, and Siemens scanners. We use the Phase A FLAIR acquisitions from 10 participants for cross-scanner harmonisation, brain-age prediction, and site classification.

\noindent\textbf{ADNI:} The Alzheimer's Disease Neuroimaging Initiative (ADNI)~\cite{jack2008adni} is a longitudinal multi-site study of Alzheimer's disease and ageing. We use T1w MRI scans from 1,042 subjects acquired on Siemens scanners and 260 subjects acquired on GE scanners to evaluate cross-vendor generalisation and Alzheimer's disease classification.

\noindent\textbf{OASIS-3:} A longitudinal neuroimaging cohort studying healthy ageing and cognitive decline~\cite{oasis}, containing multimodal MRI acquired across multiple sessions. We use 34 subjects with paired T2w and FLAIR scans and 25 subjects with paired T1w and T2w scans to evaluate out-of-distribution harmonisation.

\noindent\textbf{WMH Challenge:} The White Matter Hyperintensity (WMH) Challenge~\cite{Kuijf2019} comprises 60 clinical subjects acquired at three independent medical centres using different 3T scanner platforms: Utrecht (Philips Achieva), Amsterdam (GE Signa HDxt), and Singapore (Siemens TrioTim). We use this cohort to evaluate cross-site white matter hyperintensity segmentation.

\noindent\textbf{BraTS 2021:} The BraTS 2021 dataset~\cite{baid2021rsna} comprises multimodal MRI scans from patients with gliomas, including high- and low-grade tumours, with annotated tumour subregions. We use this dataset for qualitative assessment of pathology preservation following harmonisation.

All external datasets (except BraTS 2021, which was already skull-stripped) were preprocessed using the same registration, skull-stripping, and intensity-normalisation pipeline as the clinical cohort.

\subsection{Image preprocessing}

All three-dimensional volumes were rigidly registered to MNI152 space and skull-stripped using SynthStrip~\cite{synthstrip} unless noted otherwise in ablation experiments (Tables~\ref{tab:skull_ablation_retrieval} and~\ref{tab:skull_ablation_ssim_psnr}). For 2D MR-CLIP training, every second slice from the central 100 slices of each volume was retained. Acquisition plane (axial, coronal or sagittal) was inferred from voxel spacing by selecting the axis with highest in-plane resolution; isotropic volumes defaulted to axial sampling. Slices were resized to $224\times 224$ pixels and intensity-normalized to $[0,1]$. For harmonisation training, the same subsampling strategy was used whereas all slices were used at test time.

\subsection{MR-CLIP: Metadata-guided Contrastive Learning}

MR-CLIP learns contrast-aware representations by aligning MRI images with structured natural-language prompts derived from their DICOM acquisition metadata in a shared embedding space as shown in Fig. \ref{fig:overview}b ~\cite{avci2025mrclipefficientmetadataguidedlearning,avci2025metadataaligned3dmrirepresentations}. A 2D vision encoder (ViT-B/16) embeds individual slices and a 3D volumetric ViT variant encodes full volumes; with the text encoder embedding the metadata. Acquisition parameters are first converted into natural-language prompts using a fixed sentence template (``A brain MRI, plane \ldots, Scanner (Manufacturer, Model, Field Strength): (\ldots), Acquisition (Description, Sequence, Variant): (\ldots), Imaging Parameters (Echo Time, Repetition Time, Inversion Time, Flip Angle): (\ldots)'') and projected by a paired text transformer into the shared space.

\textbf{Metadata grouping and contrast-aware labels.}\label{contrast-group}
A key challenge in MRI is that minor acquisition parameter changes (e.g.\ small differences in TE or TR) do not meaningfully alter image contrast, so naively treating each unique parameter combination as a distinct class would over-fragment the label space. We address this by constructing semantically consistent \emph{contrast-aware groups}: categorical DICOM tags (manufacturer, scanner model, imaging plane, field strength, sequence type, sequence variant, flip angle) are grouped by exact value, while numerical timing parameters are discretised where TE and TR are jointly binned on a $20\times 20$ grid spanning the ranges observed in the training cohort, and TI is binned separately (more than 85\% of scans lack an inversion pulse) as shown in Fig. \ref{fig:supp_data}b. Series description is excluded from group construction due to site-dependent variability and used only as auxiliary text in the prompt. Each unique combination of grouped categorical and discretised numerical values defines one contrast-aware supervisory label. Alternative discretisations were explored in ablations and a full comparison is given in Table~\ref{tab:grid_transfer}; exemplar prompt descriptions are in Table~\ref{tab:top_10_metadata_verbatim}.

\textbf{Supervised contrastive objective.}
Standard CLIP uses one-to-one image--text pairs with an InfoNCE loss~\cite{infonce}, which would incorrectly penalise near-identical acquisitions that differ only in minor parameter values. Instead, MR-CLIP uses a bidirectional Supervised Contrastive (SupCon) loss~\cite{supcon} that treats all samples sharing the same contrast-aware group label as positives for each anchor, a many-to-many formulation naturally suited to the heterogeneity of clinical MRI data. Crucially, this formulation also promotes anatomy invariance: since positive pairs within a batch include slices or volumes from different subjects and different anatomical levels that share the same acquisition parameters, the objective forces the encoder to discard anatomical and subject-specific details and focus representation learning purely on contrast-dependent features. Let $z_i$ denote the $L_2$-normalised embedding of anchor $i$ (image or text) and $P(i)$ the set of all batch embeddings sharing the same contrast label as $i$ (excluding $i$ itself); $A(i)$ denotes all other batch embeddings. For temperature $\tau$:

\begin{equation}
\mathcal{L}_{i}=-\frac{1}{|P(i)|}\sum_{p\in P(i)}\log\frac{\exp(z_{i}^{\top}z_{p}/\tau)}{\sum_{a\in A(i)}\exp(z_{i}^{\top}z_{a}/\tau)}.
\end{equation}
The final training objective symmetrises over both image-to-text and text-to-image directions:
\begin{equation}
\mathcal{L}_{\mathrm{MR\text{-}CLIP}}=\tfrac{1}{2}\left(\mathcal{L}^{\mathrm{img}\rightarrow\mathrm{text}}+\mathcal{L}^{\mathrm{text}\rightarrow\mathrm{img}}\right).
\end{equation}
By aligning slices from different subjects and anatomical locations that share the same acquisition protocol, MR-CLIP learns contrast representations that are invariant to anatomy and subject identity rather than merely memorising image--text pairs.

\textbf{Artifact detection head.} To enable artifact detection within MR-CLIP, an auxiliary classification head, implemented as a MLP, is attached to the unprojected visual representations generated by the 3D Vision Transformer (ViT) encoder. The artifact head predicts across six categories: clean, motion, noise, ghosting, bias field, and Gibbs ringing.

Artifact supervision is provided during training via online synthetic corruption. For each input volume, corruption is applied with probability $p = 0.5$. If selected, one of the five artifact types is uniformly sampled and applied via the corresponding TorchIO transformation~\cite{perez-garcia_torchio_2021}; otherwise, the volume remains uncorrupted and is assigned to the clean class. Because transformations are applied on the fly, the ground-truth artifact label $y_{\mathrm{art}} \in \{1, \dots, 6\}$ is directly determined by the sampling step.

The overall training objective combines the primary multimodal alignment loss with artifact classification:
\begin{equation}
    \mathcal{L}_{\mathrm{total}} = \mathcal{L}_{\mathrm{MR-CLIP}} + \lambda_{\mathrm{art}} \mathcal{L}_{\mathrm{CE}}\left(h_{\mathrm{art}}(\mathbf{z}_{\mathrm{raw}}),\, y_{\mathrm{art}}\right),
\end{equation}
where $\mathcal{L}_{\mathrm{CE}}$ is the standard cross-entropy loss, $h_{\mathrm{art}}(\cdot)$ represents the artifact classification head operating on the unprojected visual representation $\mathbf{z}_{\mathrm{raw}}$, and $\lambda_{\mathrm{art}} = 1.0$ controls the auxiliary loss weight.

At inference, the image–metadata cosine similarity and the probability assigned to the clean class by the artifact head provide complementary measures for quality control. The former reflects the consistency between the image representation and its acquisition metadata, whereas the latter reflects the model's confidence that the volume is free of detectable artifacts. 

\subsection{DIST-CLIP: Anatomy--Contrast Disentanglement and Harmonisation}

The core objective of DIST-CLIP is to translate a source MRI scan into any target acquisition domain, defined either by a reference image or by DICOM metadata, while strictly preserving the underlying anatomical structure. To achieve this, the framework explicitly factorises each scan into two complementary, disentangled representations: an anatomical map $\beta$ that captures anatomical content, and a contrast code $\theta$ obtained from a frozen MR-CLIP encoder. As illustrated in Fig.~\ref{fig:overview}c, the Anatomy Mapper extracts the structural content of the scan, while the frozen MR-CLIP encoders capture contrast information from either images or metadata. These representations are then fused in the Style Fusion Decoder (SFD) to synthesise the final harmonised image, allowing anatomy to remain fixed while appearance is controlled by the target contrast.

\textbf{Anatomy Mapper.} The first stage aims to isolate the underlying anatomical structure from imaging physics related variations. The source image $I_{\mathrm{src}}$ is projected by the Anatomy Mapper which is a U-Net with Instance Normalisation applied in the first layer into a structural representation $\beta_{\mathrm{src}}$. This representation captures the brain's morphological geometry, such as tissue boundaries, sulcal patterns and global anatomical layout, while remaining largely invariant to intensity and contrast differences. Importantly, because different MRI sequences emphasise different tissue properties, the mapper is designed to preserve subtle, biologically meaningful variations rather than enforcing strict cross-contrast uniformity. The resulting representation $\beta_{\mathrm{src}}$ provides a contrast-agnostic structural foundation for the decoder, onto which the style module injects the target contrast embedding $\theta$.

\textbf{Contrast encoding via MR-CLIP.} We leverage pre-trained MR-CLIP encoders to model image contrast, projecting both images and DICOM metadata into a shared contrast-aware embedding space. This allows DIST-CLIP to accept two forms of guidance: a target image, processed by the image encoder ($E_I$) to produce a visual style embedding $\theta_i = E_I(I_{\mathrm{tgt}})$, or target metadata, processed by the metadata encoder ($E_M$) to produce a textual style embedding $\theta_m = E_M(M_{\mathrm{tgt}})$. To ensure the model interprets both modalities consistently, at each training iteration we randomly condition the decoder on $\theta_i$ or $\theta_m$ (with probability $0.5$). Thus, at inference time, DIST-CLIP can be guided by either an image (DIST-CLIP/I) or metadata (DIST-CLIP/T) alone, since $\theta_i$ and $\theta_m$ were pre-trained to be aligned in a joint latent space. To further promote disentanglement and prevent the model from memorising spatial style coordinates, the style code $\theta$ is extracted from a randomly sampled slice of the target volume ensuring the style representation remains spatially invariant.

\textbf{Style Fusion Decoder and harmonisation.} The final harmonisation is performed by the SFD (Supplementary Fig.~\ref{fig:supp_sfd}), which synthesises the output image by fusing the target style embedding $\theta$ with the anatomical representation $\beta_{\mathrm{src}}$. Architecturally, the SFD is a U-Net decoder with twice the channel width of the Anatomy Mapper to support rich feature modulation. Skip connections from the Anatomy Mapper are merged at each upsampling stage to preserve fine-grained spatial structure, and the target contrast is injected at every decoding layer through an Adaptive Style Transfer (AST) module. 

Within each AST block, a Multi-Layer Perceptron (MLP) maps the contrast embedding $\theta \in \mathbb{R}^{512}$ to a style query vector $Q = \text{MLP}(\theta) \in \mathbb{R}^{C}$. The keys $K$ and values $V$ are projected from the spatial feature maps $F \in \mathbb{R}^{H \times W \times C}$ of the previous layer. The cross-attention output $A = \text{Softmax}(Q K^T / \sqrt{C}) V$ dynamically modulates the feature map based on the target acquisition style. A bottleneck AST module uses 8-head attention, while upsampling AST modules use lighter linear-attention variants for computational efficiency. The attention output is then modulated using channel-wise scale $\gamma$ and shift $\beta$ parameters predicted by a parallel MLP from $\theta$, driving the spatial Adaptive Instance Normalisation (AdaIN)~\cite{huang2017arbitrarystyletransferrealtime}:
\begin{equation}
\text{AdaIN}(A, \gamma, \beta) = \gamma \cdot \left(\frac{A - \mu(A)}{\sigma(A)}\right) + \beta
\end{equation}
where $\mu(A)$ and $\sigma(A)$ represent the spatial mean and standard deviation of $A$ per channel. This progressive, layer-wise modulation allows both global contrast and fine texture to be controlled while keeping the anatomical structure anchored to $\beta_{\mathrm{src}}$.

\textbf{Training objectives.}
DIST-CLIP is trained with a composite objective that jointly enforces anatomical preservation, high-fidelity reconstruction, and MR-CLIP-guided style consistency:
\begin{equation}
\begin{aligned}
\mathcal{L}_{\mathrm{total}} = {} & \lambda_{\mathrm{rec}}\mathcal{L}_{\mathrm{rec}} + \lambda_{\mathrm{perc}}\mathcal{L}_{\mathrm{perc}}  + \lambda_{\mathrm{adv}}\mathcal{L}_{\mathrm{adv}} + \\
& \lambda_{\beta}\mathcal{L}_{\beta} + \lambda_{\mathrm{global}}\mathcal{L}_{\mathrm{global}} + \lambda_{\mathrm{dir}}\mathcal{L}_{\mathrm{dir}}
\end{aligned}
\end{equation}
where  $\mathcal{L}_{\mathrm{rec}}$ is an $\ell_1$ reconstruction term between the synthesised image $I_{\mathrm{rec}}$ and target $I_{\mathrm{tgt}}$.
$\mathcal{L}_{\mathrm{perc}}$ is a multi-layer VGG-19 perceptual loss.
$\mathcal{L}_{\mathrm{adv}}$ is a patch-based adversarial term that encourages synthesised slices to be locally indistinguishable from real target contrast images.
We use a 2D PatchGAN discriminator implemented in MONAI~\cite{monai2020}, following the Pix2PixHD architecture~\cite{wang2018pix2pixhd}.

To preserve fine-grained anatomical structure while promoting contrast invariance, we adopt the patch-level contrastive alignment loss ($\mathcal{L}_{\mathrm{\beta}}$) from HACA3~\cite{haca3}. Rather than enforcing identical features across contrasts, which can be overly restrictive, this loss aligns corresponding patches in the source ($\beta_{src}$) and target ($\beta_{tgt}$) anatomy maps while separating non-corresponding patches. Implemented as a patch-wise InfoNCE objective~\cite{infonce}, it encourages structural consistency across contrasts while allowing necessary feature variations. Formally, the loss is defined as:
\begin{equation}
\mathcal{L}_{\beta} = \mathbb{E}_{\beta_{src}, \beta_{tgt}} \sum_{i \in P} \left[ - \log \frac{e^{\text{sim}(z_i, z'_i) / \tau}}{\sum_{k \in P} e^{\text{sim}(z_i, z'_k) / \tau}} \right]
\end{equation}
where $P$ is the set of all patch locations. For each source patch $z_i$ in $\beta_{src}$, $z'_i$ is the corresponding patch in $\beta_{tgt}$, while $z'_k$ (for all $k \in P$) represents \emph{all} patches in the target map, $\text{sim}(\cdot, \cdot)$ is the cosine similarity, and $\tau$ is the temperature parameter.


Global style consistency was enforced by
\begin{equation}
    \mathcal{L}_{\mathrm{global}}=\tfrac{1}{2}(1-\cos(E_I(I_{\mathrm{rec}}),\theta_i))+\tfrac{1}{2}(1-\cos(E_I(I_{\mathrm{rec}}),\theta_m))
\end{equation}
Directional consistency followed StyleGAN-NADA~\cite{gal2021stylegannada}:
\begin{equation}
\mathcal{L}_{\mathrm{dir}}=1-(\Delta I\cdot\Delta M)/(\|\Delta I\|\|\Delta M\|)
\end{equation}
with $\Delta I=E_I(I_{\mathrm{rec}})-E_I(I_{\mathrm{src}})$ and $\Delta M=E_M(M_{\mathrm{tgt}})-E_M(M_{\mathrm{src}})$.
Loss weights were selected to keep all objectives on a comparable scale, ensuring balanced optimization and preventing any single term from dominating training: $\lambda_{\beta}=0.1$, $\lambda_{\mathrm{rec}}=10$, $\lambda_{\mathrm{perc}}=\lambda_{\mathrm{adv}}=1$, and $\lambda_{\mathrm{global}}=\lambda_{\mathrm{dir}}=1$.

\textbf{DIST-CLIP-2.5D.}
For volumetric harmonisation, we trained a separate DIST-CLIP-2.5D model without the Anatomy Mapper due to memory constraints. Contrast conditioning was provided by the frozen three-dimensional MR-CLIP image encoder: the contrast representation $\theta$ was computed once from the target volume and broadcast to all slices during decoding, enforcing consistent contrast across the volume. The training objectives and data partitions were otherwise matched to those used for slice-wise DIST-CLIP/I, except that $\mathcal{L}_{\beta}$ was omitted. The two models differed only in their MR-CLIP contrast encoder (3D versus 2D) and harmonisation network weights.

\subsection{MR-SYNTH: Metadata-Conditioned Image Synthesis}

MR-SYNTH is a 3D Latent Diffusion Model (LDM)~\cite{rombach2022high} that synthesises full-resolution brain MR volumes conditioned on acquisition metadata. The model builds upon the MAISI-v2 pretrained 3D diffusion backbone~\cite{zhao2026maisi} and operates entirely within the compressed latent space of a frozen KL-variational autoencoder. At inference, a structured metadata prompt specifying contrast weighting, sequence timings ($\mathrm{TE}$, $\mathrm{TR}$, $\mathrm{TI}$), flip angle, and scanner hardware is encoded by MR-CLIP's text encoder and used to steer the generative process, producing anatomically coherent volumetric outputs whose appearance reflects the requested acquisition protocol.

\textbf{Metadata conditioning.}
Rather than conditioning on discrete modality labels, MR-SYNTH accepts continuous, semantically structured MR-CLIP text embeddings as its conditioning signal. A lightweight two-layer projection MLP maps each $L_2$-normalised 512-dimensional metadata embedding  into the UNet's time-embedding space, where it is added to the time embedding before being broadcast into every residual block. This additive injection requires no architectural modification to the pretrained backbone, only the projection MLP and a learnable null embedding $\varnothing$ are introduced as new parameters. At inference, the conditional and unconditional velocity predictions are interpolated via a scalar guidance scale $s$, allowing continuous control over how strongly the output adheres to the requested metadata.

\textbf{Rectified flow matching objective.}
MR-SYNTH is trained with a rectified flow matching objective~\cite{liu2022flow}. Rather than predicting noise as in standard DDPM~\cite{ho2020denoising}, the model learns to predict the velocity along a linear interpolation between a Gaussian noise sample $\mathbf{x}_0$ and a clean latent $\mathbf{x}_1$:
\begin{equation}
\mathbf{x}_t = (1-t)\mathbf{x}_0 + t\mathbf{x}_1, \quad t \in [0,1].
\end{equation}
The target velocity is $\mathbf{x}_1-\mathbf{x}_0$, and the network is trained to predict this velocity using an $L_1$ loss. The resulting rectified flow enables efficient numerical integration with substantially fewer sampling steps than conventional DDPM sampling, which we exploit during inference. 


\paragraph{Architectures and Training}

\textbf{MR-CLIP.}
The default image encoder is a Vision Transformer (ViT-B/16); metadata were encoded with the paired text transformer. Image and metadata embeddings were $L_2$-normalised to unit length before contrastive learning. We additionally trained a three-dimensional volumetric ViT variant and a 2.5D variant that aggregated slice-level predictions for volume-level retrieval and classification. Models were implemented in PyTorch. Two-dimensional MR-CLIP was trained on three NVIDIA A100 GPUs (40\,GB) with per-GPU batch size 3,000 using gradient accumulation and sharded contrastive loss following OpenCLIP~\cite{open_clip}. Three-dimensional MR-CLIP used per-GPU batch size 150 with the same sharded-loss strategy. Optimisation used Adam~\cite{kingma2017adammethodstochasticoptimization} ($\mathrm{lr}=10^{-4}$, $\beta_1=0.9$, $\beta_2=0.98$, weight decay $0.2$) for 100 epochs with 2,000 warm-up steps. Gradient checkpointing was enabled. Training augmentations included random affine transforms, random resized crops, Gaussian blur and horizontal flips; patch dropout (0.5) and text dropout (0.2) were applied to the image and text towers respectively.

\textbf{DIST-CLIP.}
DIST-CLIP was trained for 25 epochs on NVIDIA A100 GPUs using Adam ($\mathrm{lr}=10^{-4}$, batch size 25) with MR-CLIP encoders frozen throughout. The anatomy mapper, style fusion decoder and discriminator were optimised jointly end-to-end. For each training iteration, we sampled a subject with at least two available contrasts and constructed an ordered source–target contrast pair, yielding the corresponding paired volumes.

\textbf{MR-SYNTH.}
The 3D UNet backbone has channel widths $[64, 128, 256, 512]$, two residual blocks per resolution level, and Flash Attention at the two deepest levels. The original discrete modality-label embedding is replaced by a two-layer projection MLP that additively injects 512-dimensional $L_2$-normalised MR-CLIP text embeddings into the time-embedding stream; a learnable null embedding supports classifier-free guidance at inference. Starting from the pretrained MAISI-v2 rectified flow brain checkpoint, training proceeds in two phases of 100 epochs each: \textbf{Phase~1} freezes the backbone and optimises only the projection MLP and null embedding ($\mathrm{lr}=10^{-3}$); \textbf{Phase~2} fine-tunes the full model with a differential learning rate, lower for the backbone ($\mathrm{lr}=10^{-5}$) and higher for the projection head ($\mathrm{lr}=10^{-4}$), to balance adaptation and stability. Both phases use Adam optimisation with polynomial learning-rate decay and mixed-precision training, conducted on Quadro RTX 8000 GPU. At inference, generation requires 30 denoising steps with CFG scale $s=6.0$.

\subsection{Evaluation protocols}

\textbf{Anatomical representation analysis.}
Anatomical maps were extracted with the trained DIST-CLIP anatomy mapper. Segmentation consistency was evaluated on paired cross-contrast subjects from the harmonisation cohort ($n=53$ subjects with at least three registered acquisitions of distinct contrasts, including T1w, T2w, FLAIR, PD and T2*). For each acquisition we segmented both the native scans used in training and the corresponding anatomical $\beta$-maps.

Tissue maps were generated independently for every image using the SPM12 unified segmentation method~\cite{ashburner2005unified}, yielding discrete labels for grey matter (label 1), white matter (label 2) and cerebrospinal fluid (label 3). We deliberately used SPM12 because it is a well-established, non-deep-learning statistical segmentation method, avoiding the additional learned-model biases that could arise from applying a neural-network-based segmentation model to the non-conventional intensity distributions of the $\beta$-maps. Tissue volume (ml) was computed as the number of labelled voxels multiplied by voxel volume. For each subject and tissue class, we computed the coefficient of variation ($\mathrm{CV}=\sigma/\mu$) of tissue volumes across all available acquisitions, separately for the raw images and $\beta$-maps. Lower CV indicates more stable tissue quantification across contrasts.

For the ADNI cross-vendor classification experiment, models were trained using the Siemens cohort ($n=1,042$; 116 AD and 926 CN) and evaluated on the independent GE cohort ($n=260$; 39 AD and 221 CN). Five-fold cross-validation was performed using subject-disjoint splits within the Siemens cohort, such that all acquisitions from a given subject were assigned to the same fold. Five separate models were trained, with each fold serving once as the held-out validation set, and performance was averaged across the five folds. Each of the five trained models was subsequently evaluated on the subject-disjoint GE cohort to assess cross-vendor generalisation, and the GE performance was likewise averaged across the five models. Models were trained separately using either raw T1-weighted MRI or $\beta$-maps as input, and balanced accuracy was used as the evaluation metric. The classifier is a standard ResNet-18 backbone trained on 3D volumes with cross-entropy loss, learning rate $10^{-4}$, AdamW optimizer, and early stopping on validation loss.

\textbf{Harmonisation metrics.}
Image fidelity was quantified using peak signal-to-noise ratio (PSNR) and structural similarity index (SSIM) against target slices on the held-out test pairs, reported for every available source--target contrast direction (Supplementary Fig.~\ref{fig:supp_metrics}). To assess image-based conditioning (DIST-CLIP/I and HACA3) fairly, we use a reference image from a different subject acquired under the same target protocol to extract the target style, while evaluation is performed against the corresponding registered ground-truth target scan of the source subject. PSNR and SSIM were computed in intensity space on $224\times 224$ slices after aligning synthesised outputs to target contrast. Volumetric consistency was assessed qualitatively by inspecting axial, sagittal and coronal reformations (Supplementary Fig.~\ref{fig:supp_visual}). For T1w$\rightarrow$T2w harmonisation, FreeSurfer SynthSeg~\cite{synthseg} was applied to each method's harmonised NIfTI and to the reference T2w scan; per case we computed mean Dice across labelled structures between reference and harmonised parcellations and averaged across cases (held-out clinical set, $n=20$; OASIS-3 test set, $n=14$).

\textbf{Downstream analysis on ON-Harmony data.}
 On ON-Harmony~\cite{warrington2025onharmony} ($n=10$ participants), we evaluated whether harmonisation suppresses scanner-specific signatures while preserving biological signal using brain-age regression and scanner-vendor classification on harmonised versus native slices. We used a 4-layer 3D CNN for the brain-age regression model (evaluated via mean absolute error, MAE). For the site classification task, we employed a 3D ResNet-18 to predict the acquisition site. All models were trained and evaluated using a leave-one-subject-out cross-validation.

\textbf{Downstream segmentation analysis on WMH data.}
One nnU-Net~\cite{isensee2021nnu} 3D full-resolution segmentation model was trained per site on its own 20 subjects, yielding three site-specific segmenters. For each cross-site direction (e.g.\ Amsterdam $\rightarrow$ Utrecht), the source subjects
were harmonised by each method to match the target site contrast; the resulting volumes were then fed to the target-site segmenter, and the predicted masks were compared
against the original source-site ground-truth annotations using  voxel-level Dice Similarity Coefficient (DSC).

\textbf{Baselines.}
MR-CLIP was compared against frozen and fine-tuned BiomedCLIP~\cite{biomedclip}, ViT-S/16 and ViT-B/16 models trained with standard InfoNCE loss~\cite{infonce}, and a supervised 3D ResNet for few-shot sequence classification. Harmonisation was compared against HACA3~\cite{haca3} and TUMSyn~\cite{tumsyn}, including slice-wise DIST-CLIP/I, DIST-CLIP/T and volumetric DIST-CLIP-2.5D, under the same preprocessing and held-out test partitions. HACA3 and TUMSyn were run using their public implementations and original weights with recommended settings. HACA3 was run with skull and skull stripping is applied during evaluation. All methods received identical registration, skull-stripping, resampling before metric computation.

\textbf{Cross-modal retrieval and linear probing.}
We evaluated recall at ranks 1, 5 and 10 (R@1/5/10) for slice-to-metadata, metadata-to-slice (volume), and volume-to-metadata retrieval on the held-out test set. A retrieval was counted as correct if the retrieved image or metadata had the correct contrast label (out of 1,415 contrast groups), as the task is to identify the correct contrast regardless of anatomy. Volume-level image embeddings were obtained by majority vote across slice embeddings (2D/2.5D) or direct volumetric encoding (3D). Linear probes were trained on frozen image embeddings to predict grouped metadata labels and sequence class.

\textbf{Few-shot sequence classification.}
A linear classifier was trained on frozen MR-CLIP image embeddings with $\{1,2,4,8,16,32,64,128,256,512\}$ labelled volumes per sequence class and compared against a supervised 3D ResNet trained end-to-end on the same labelled subsets.

\textbf{Metadata quality control.}
On held-out test scans, we introduced synthetic metadata corruptions~\cite{avci2025metadataaligned3dmrirepresentations} at rates of 0, 25, 50 and 75\% of the test set. Corruptions comprised: (i)~\textit{Numerical error}: small scaling of TE/TR/TI within range, medium shifts outside expected sequence ranges, or large unit errors (for example, seconds versus milliseconds); (ii)~\textit{Wrong tag}: incorrect series description, sequence fields, and scanner identifiers; and (iii)~\textit{Missing tag}: progressively removing series description, sequence information, and scanner fields. Mean cosine similarity between $E_I(I)$ and $E_M(M)$ was computed as a function of corruption rate. Detection performance was summarised by the area under the receiver operating characteristic curve (AUC) at 50\% corruption rate, using one minus cosine similarity as an anomaly score.

\textbf{Statistical tests and units.} All statistical comparisons were performed using a two-sided Wilcoxon signed-rank test, except for the sequence-classification comparison in Fig.~\ref{fig:metadata}d, for which McNemar's test was used. The statistical unit was the subject except harmonisation SSIM/PSNR experiments, where the unit is one sided paired contrast translation. When results were available at the slice level, slice-level measurements were first averaged within each subject before statistical testing. Similarly, repeated acquisitions or multiple observations from the same subject were averaged before statistical tests. All reported comparisons were therefore performed on paired subject-level measurements, using a significance threshold of $P<0.05$

\subsection{Software and reproducibility}

Code and pre-trained model weights for MR-CLIP, DIST-CLIP and evaluation scripts are available in the MaRaI repository (\url{https://github.com/myigitavci/MaRaI}). Experiments were implemented in PyTorch; software versions, random seeds, full hyperparameter tables, and downstream model architectures are documented in the repository configuration files.

\backmatter







\section*{Declarations}

\subsection*{Funding}
This work was supported by the UK Engineering and Physical Sciences Research Council (EPSRC) Centre for Doctoral Training in Data-Driven Health (DRIVE-Health; grant EP/Y035216/1) at King's College London, with additional support from deepc GmbH and the Scientific and Technological Research Council of T\"urkiye (TUBITAK) 2213-A Overseas Graduate Scholarship.
\subsection*{Competing interests}
M.Y. Avci's PhD is partially funded by deepc GmbH. The other authors declare no competing interests.

\subsection*{Ethics approval and consent to participate}
Approved under UK Health Research Authority generic approval (IRAS ID 349531; REC reference 24/ES/0099).

\subsection*{Data availability}
Clinical MRI data from King's College Hospital and Guy's and St Thomas' NHS Foundation Trust are not publicly available owing to governance restrictions. ON-Harmony is available on OpenNeuro (\texttt{ds004712}). WMH Challenge, BrATS 2021, ADNI and OASIS-3 are available from their respective repositories subject to data-use agreements.

\subsection*{Code availability}
Code and pre-trained weights are available at \url{https://github.com/myigitavci/MaRaI}.

\subsection*{Author contributions}
M.Y.A., P.B., V.F., N.G., P.W., M.Y., S.O. and J.C. designed the study. M.Y.A., P.B., V.F., and J.C. developed methodology and software. M.Y.A., P.B., and N.G. performed experiments and analysis. M.Y.A. and J.C. wrote the manuscript with input from all authors. S.O. and J.C. supervised the project.

\begin{appendices}

\setcounter{figure}{0}
\setcounter{table}{0}
\renewcommand{\thefigure}{S\arabic{figure}}
\renewcommand{\theHfigure}{S.\arabic{figure}}
\renewcommand{\thetable}{S\arabic{table}}
\renewcommand{\theHtable}{S.\arabic{table}}

\section{Supplementary information}\label{sec:supplementary}

\subsection{Metadata distribution}

\begin{figure}[p]
    \centering
    \includegraphics[width=0.99\textwidth]{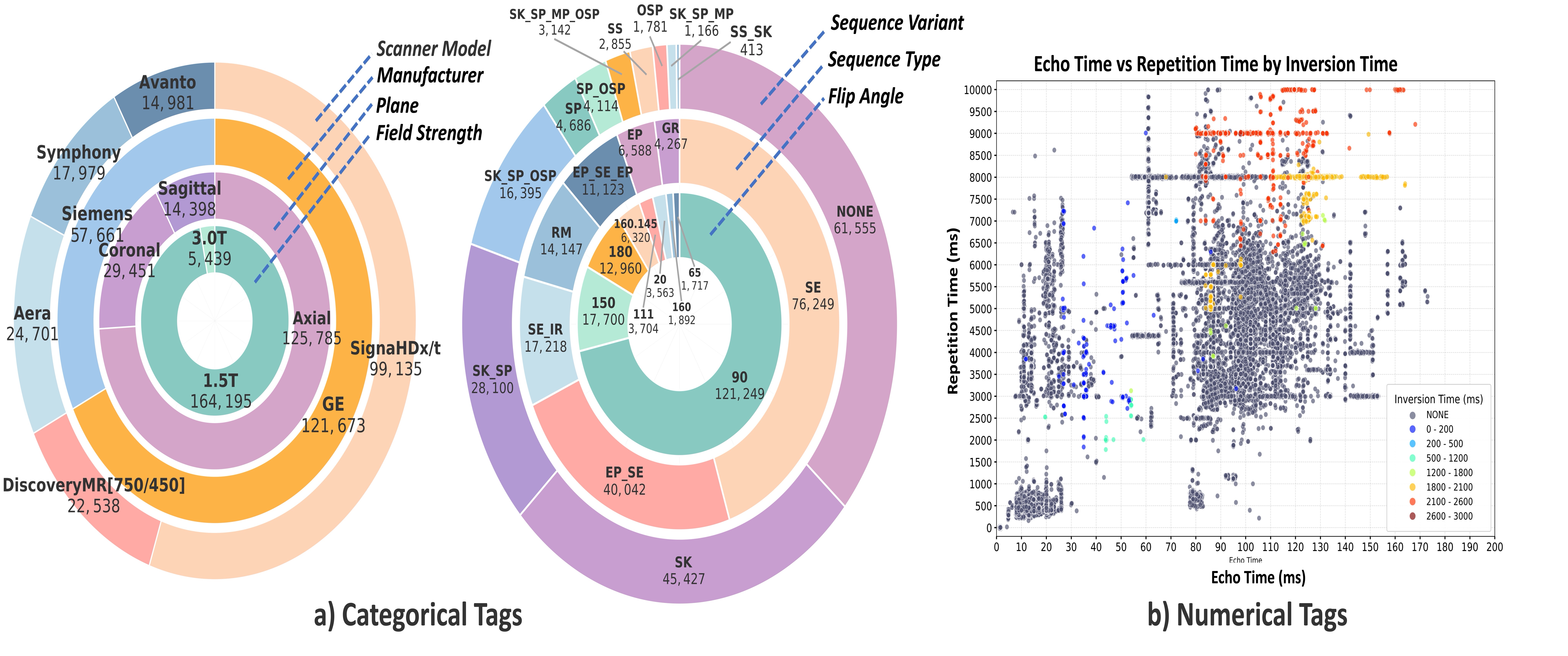}
    \caption{
    \textbf{Metadata distribution in the clinical cohort.}
    \textbf{a,}
    Categorical tags: nested distributions of imaging plane, field strength, manufacturer, scanner model, flip angle, sequence type and sequence variant.
    \textbf{b,}
    Numerical tags: echo time (TE) versus repetition time (TR), with inversion time (TI) indicated by colour.
    }
    \label{fig:supp_data}
\end{figure}

Supplementary Fig.~\ref{fig:supp_data} summarises DICOM metadata distribution in the dataset.

\begin{figure}[p]
\centering
\includegraphics[width=0.99\textwidth]{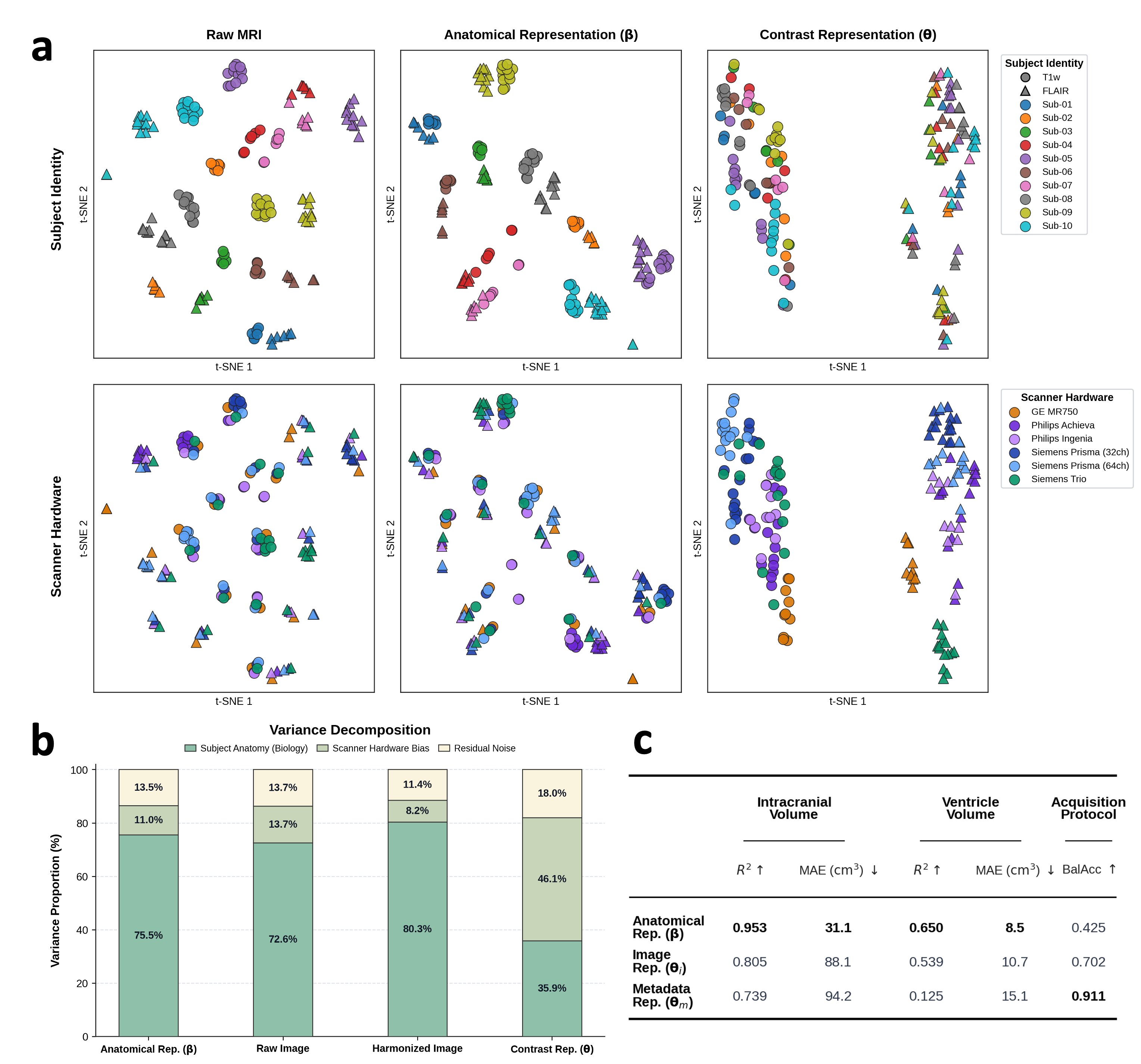}
\caption{\textbf{Disentanglement of anatomical and acquisition-related information in $\beta$ and $\theta$.}
\textbf{a,} t-SNE projections of raw MRI, the anatomical representation ($\beta$), and the contrast representation ($\theta$) for the ON-Harmony travelling-head cohort, colour-coded by subject identity (top row) and by scanner hardware (bottom row); marker shape indicates sequence (T1w: circle, FLAIR: triangle). \textbf{b,} Variance decomposition of subject-, scanner-, and residual-attributable components for $\beta$, raw images, harmonised reconstructions, and $\theta$, estimated using a linear mixed-effects model on the ON-Harmony travelling-head cohort. \textbf{c,} Linear probing results on the ADNI cohort, showing $R^2$ and mean absolute error (MAE) for intracranial and ventricular volume, and balanced accuracy (BalAcc) for acquisition-protocol classification, from $\beta$, the image contrast representation ($\theta_i$), and the metadata representation ($\theta_m$).}
\label{fig:beta_disentanglement}
\end{figure}

\subsection{Disentanglement of anatomical and acquisition-related information}

A central question for the proposed factorisation is whether $\beta$ captures genuine anatomical structure and whether $\beta$ and $\theta$ encode distinguishable information. We addressed this using complementary analyses on the multi-site travelling-head ON-Harmony cohort (Fig.~\ref{fig:beta_disentanglement}a--b) and the ADNI cohort (Fig.~\ref{fig:beta_disentanglement}c).

In low-dimensional manifold projections of the ON-Harmony data (Fig.~\ref{fig:beta_disentanglement}a), raw MRI scans show clustering primarily by subject, with additional separation by sequence contrast. In $\beta$, scans from the same subject become more closely grouped, while separation between T1w and FLAIR scans and scanner-specific structure are reduced. In contrast, $\theta$ shows stronger organisation by acquisition characteristics, with clustering primarily by sequence contrast and scanner hardware rather than subject identity. These projections therefore suggest that $\beta$ and $\theta$ emphasise different sources of variation.

Variance decomposition on the same travelling-head cohort provides quantitative support for this observation (Fig.~\ref{fig:beta_disentanglement}b). A linear mixed-effects model attributes 75.5\% of variance in $\beta$ to subject identity and 11.0\% to scanner hardware, compared with 72.6\% and 13.7\%, respectively, in raw images. In harmonised reconstructions, the subject-attributable component increases to 80.3\%, while the scanner-attributable component decreases to 8.2\%, consistent with reduced acquisition-related variation. Conversely, 46.1\% of variance in $\theta$ is attributable to scanner hardware, indicating substantially greater retention of scanner-related information than in $\beta$.

Supervised linear probing on ADNI provides an independent assessment of the information retained by each representation (Fig.~\ref{fig:beta_disentanglement}c). $\beta$ predicts anatomical geometry substantially better than either contrast or metadata representations, achieving $R^2=0.953$ (MAE~=~31.1~cm$^3$) for intracranial volume and MAE~=~8.5~cm$^3$ for ventricular volume, compared with an intracranial-volume MAE of 88.1~cm$^3$ for $\theta_i$ and 94.2~cm$^3$ for $\theta_m$. Conversely, acquisition-protocol classification achieves a balanced accuracy of 0.425 from $\beta$, compared with 0.702 from $\theta_i$ and 0.911 from $\theta_m$. Together, these complementary analyses show that $\beta$ preferentially retains subject- and anatomy-related information, whereas $\theta$ retains substantially more acquisition-related information.

\subsection{Complete cross-contrast harmonisation metrics}

\begin{figure}[p]
\centering
\includegraphics[width=\linewidth]{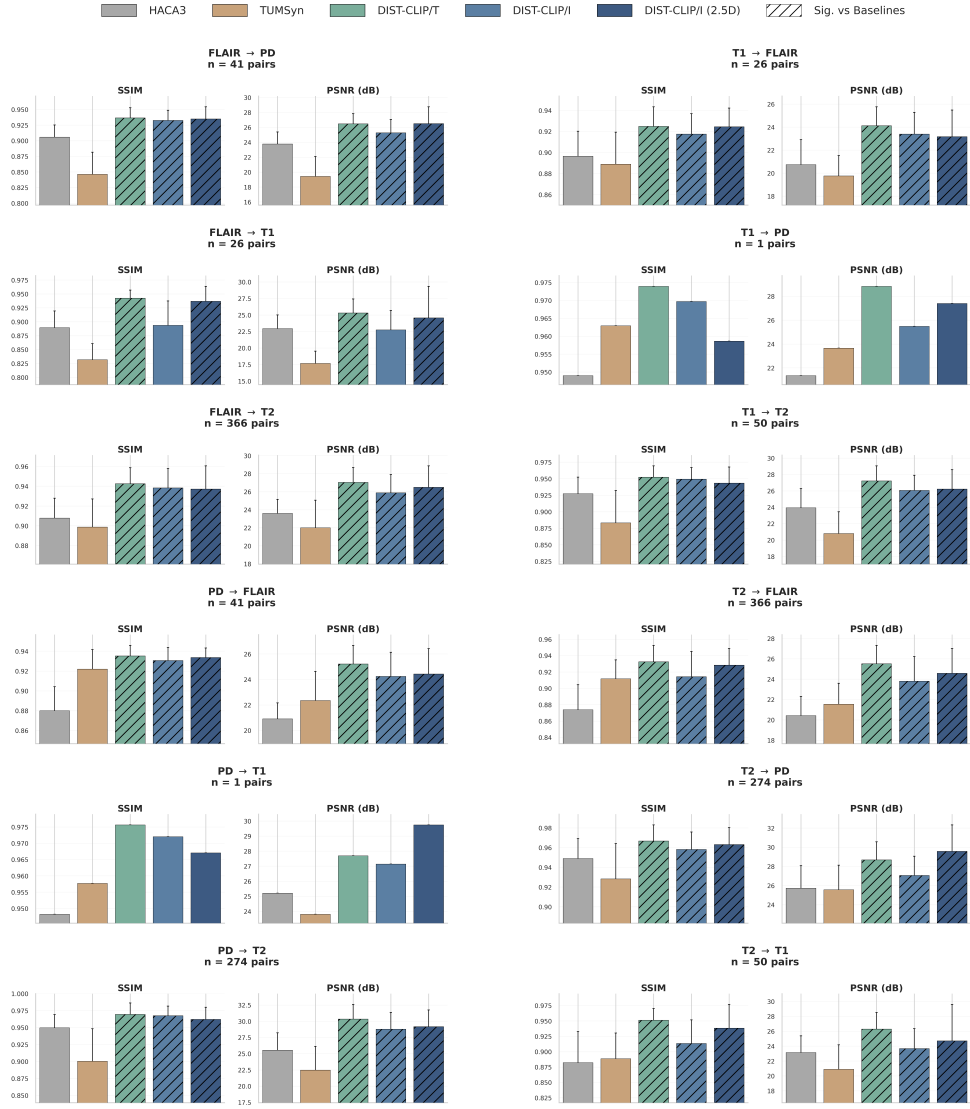}
\caption{
\textbf{Complete cross-contrast harmonisation benchmarks.}
SSIM and PSNR (dB) for all bidirectional translation tasks among T1w, T2w, PDw and FLAIR sequences. Hatched bars denote methods that achieve statistically significant improvements over both HACA3 and TUMSyn ($p<0.05$).
}
\label{fig:supp_metrics}
\end{figure}

Supplementary Fig.~\ref{fig:supp_metrics} reports mean SSIM and PSNR ($\pm$ s.d.) for every available source--target contrast pair in the harmonisation test set, comparing HACA3, TUMSyn, DIST-CLIP/T, DIST-CLIP/I and DIST-CLIP-2.5D. Panel titles indicate the translation direction and the number of one sided test pairs ($n$). Following common practice in the image synthesis literature, these metrics are computed over the entire image volume, including background regions, to facilitate direct comparison with previously published methods. Hatched bars indicate results that are statistically significantly better than both baseline methods (HACA3 and TUMSyn) ($p<0.05$).

\begin{figure}[p]
\centering
\includegraphics[width=\linewidth]{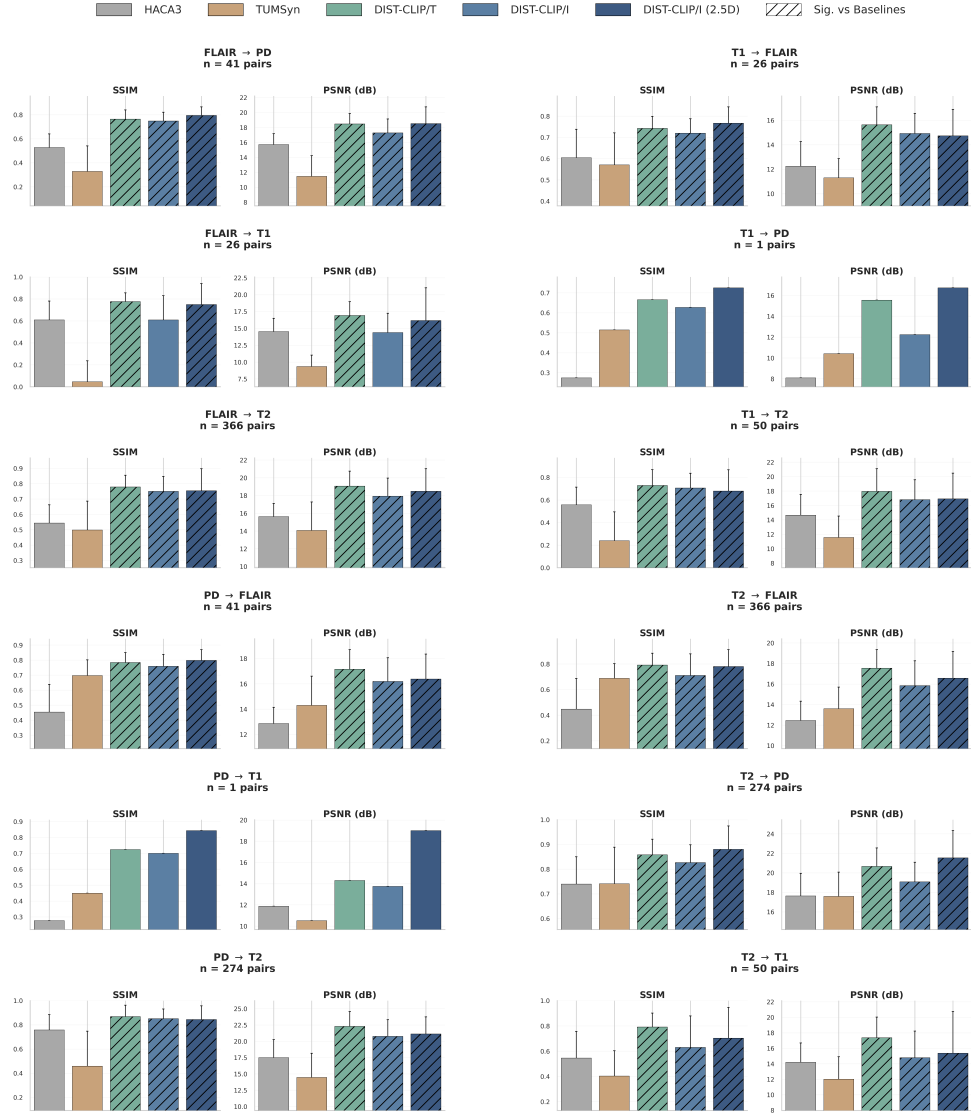}
\caption{
\textbf{Foreground-only cross-contrast harmonisation benchmarks.}
SSIM and PSNR (dB) computed exclusively within the brain foreground mask for all bidirectional translation tasks among T1w, T2w, PDw and FLAIR sequences. Hatched bars denote methods that achieve statistically significant improvements over both HACA3 and TUMSyn ($p<0.05$).
}
\label{fig:supp_metrics_masked}
\end{figure}

While most prior studies report SSIM and PSNR over the full image domain, including background voxels, these metrics can be influenced by large homogeneous regions outside the brain. To assess synthesis quality within anatomically relevant structures, we additionally computed SSIM and PSNR exclusively within the brain foreground mask. Supplementary Fig.~\ref{fig:supp_metrics_masked} presents the corresponding foreground-only evaluation for all source--target contrast pairs. 

\subsection{Volumetric consistency of harmonised outputs}

\begin{figure}[p]
    \centering
    \includegraphics[width=\linewidth]{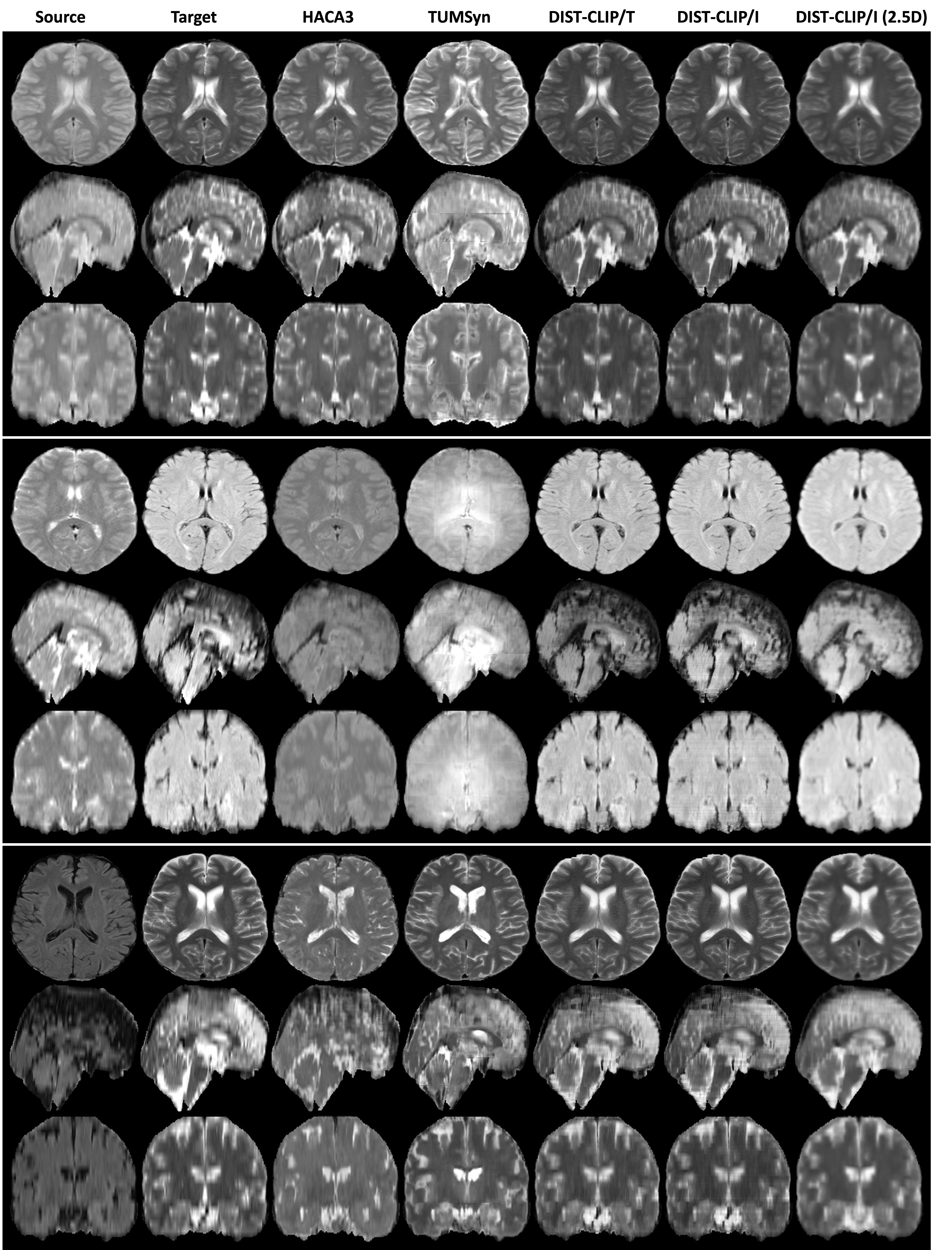}
    \caption{
    \textbf{Volumetric consistency of cross-contrast harmonisation.}
    For each target contrast block, columns show ground truth and harmonised volumes in axial, sagittal and coronal planes for representative baselines and DIST-CLIP-2.5D variant. Stair-step artefacts in orthogonal views indicate slice-wise inconsistency of purely two-dimensional synthesis; DIST-CLIP-2.5D reduces these artefacts while producing smoother intensity profiles.
    }
    \label{fig:supp_visual}
\end{figure}

Supplementary Fig.~\ref{fig:supp_visual} compares axial, sagittal and coronal views for three target contrasts (T2w, FLAIR and T1w) synthesised from a common source. Slice-independent two-dimensional harmonisation shows through-plane discontinuities in orthogonal views, whereas DIST-CLIP-2.5D yields smooth volumetric appearance at the cost of reduced high-frequency texture relative to ground truth.

\subsection{Qualitative anatomical segmentation fidelity}
\label{sec:qualitative_segmentation}

\begin{figure}[p]
\centering
\includegraphics[width=0.99\textwidth]{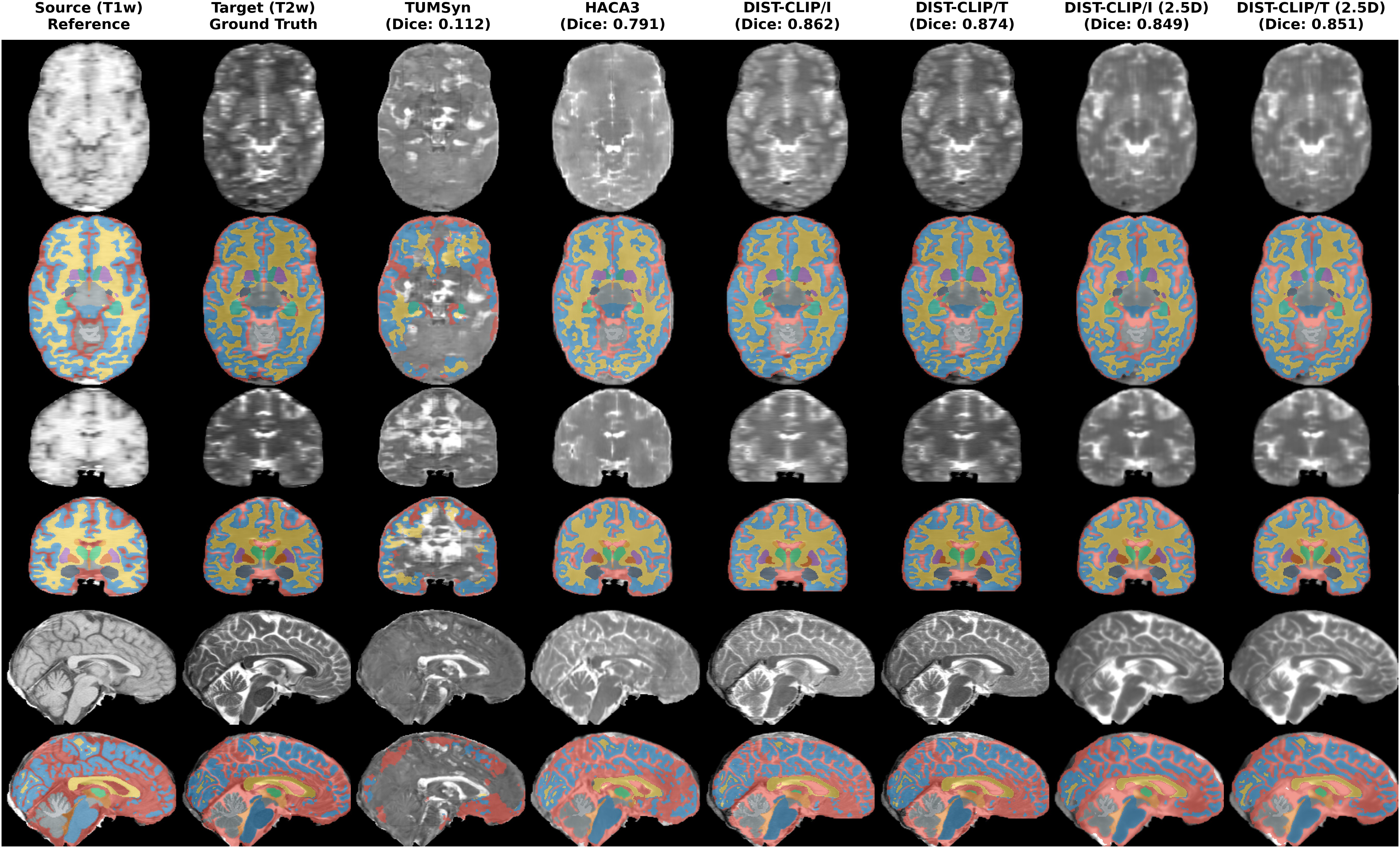}
  \caption{\textbf{Qualitative cross-modality MR harmonisation (T1w to T2w) and downstream anatomical segmentation.} 
  Visual comparison across orthogonal anatomical planes (Axial, Coronal, Sagittal) for a representative held-out unseen test subject.  
  Mean whole-brain Dice similarity coefficients relative to target ground-truth segmentations are reported above each method column. }
  \label{fig:qualitative_synthseg_overlays}
\end{figure}

Figure~\ref{fig:qualitative_synthseg_overlays} illustrates qualitative T1w-to-T2w harmonisation results alongside downstream automated SynthSeg parcellations for a representative held-out test subject. TUMSyn exhibits substantial contrast degradation and structural alterations with grid artefacts, resulting in poor downstream segmentation performance. HACA3 produces a more plausible translation but shows local over-smoothing and minor deviations in anatomical boundaries. In comparison, both the 2D and 2.5D variants of DIST-CLIP better preserve morphological detail, including sulcal--gyral topology and ventricular boundaries.

\subsection{Out-of-distribution harmonisation on ON-Harmony and OASIS-3}

We evaluated practical relevance and robustness to domain shift using cross-site harmonisation on the external ON-Harmony travelling-heads dataset~\cite{warrington2025onharmony}. To quantify whether harmonisation preserves clinically meaningful information while removing acquisition-specific variability, we trained two downstream 3D convolutional neural networks (CNNs) to classify sites and predict brain ages on the harmonised outputs of the ON-Harmony dataset. Lower site classification accuracy indicates reduced scanner- and protocol-specific information, while lower brain-age MAE indicates better preservation of anatomical structure relevant to normative ageing. An effective harmonisation method should therefore reduce site identifiability without degrading downstream predictive performance.

\begin{figure}[p]
    \centering
    \includegraphics[width=\linewidth]{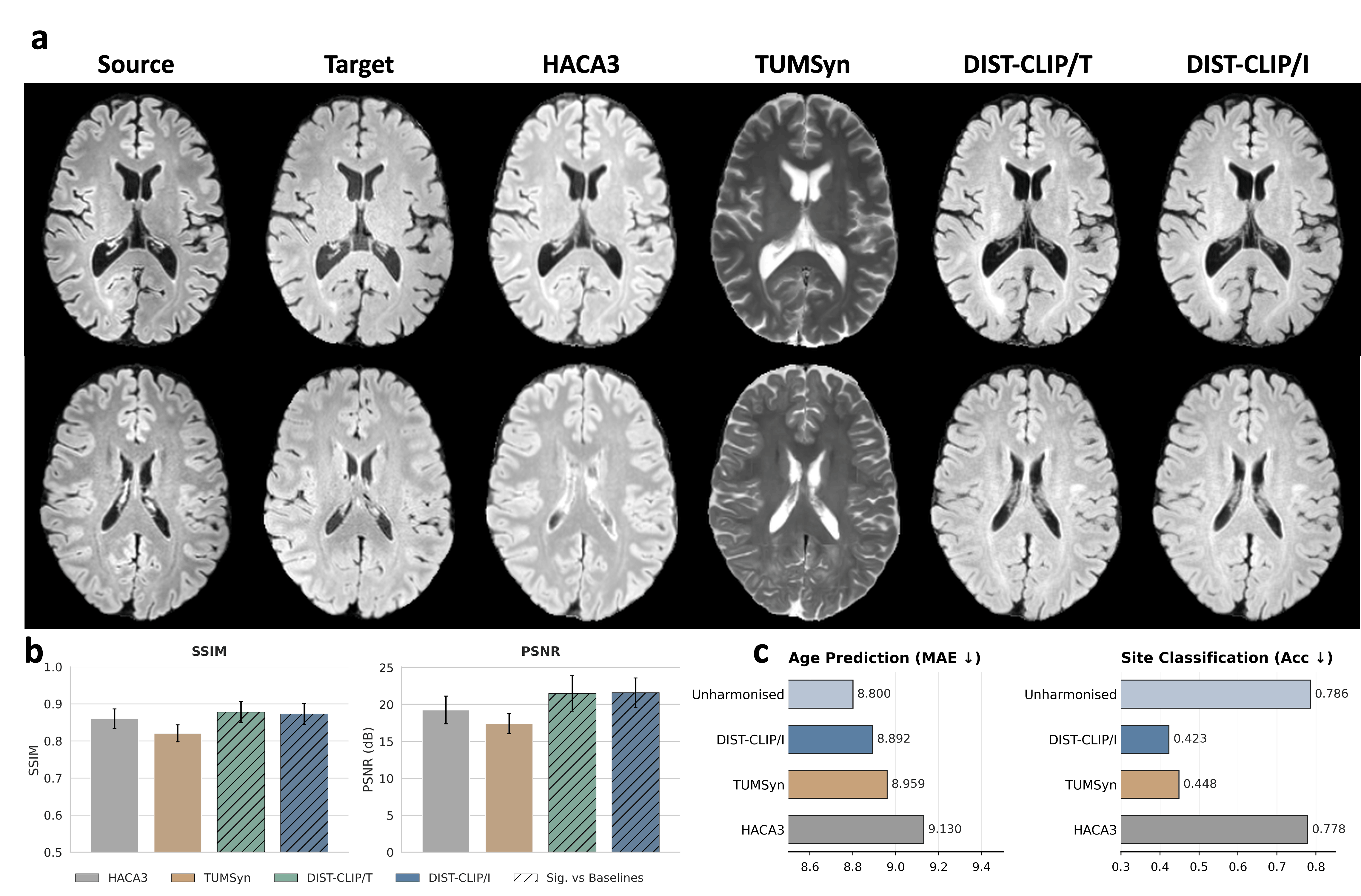}
    \caption{
        \textbf{Cross-scanner harmonisation on the external ON-Harmony travelling-head dataset ($n=10$ subjects).}
        \textbf{a,}
        Representative examples of FLAIR images harmonised across scanners while preserving anatomical structure.
        \textbf{b,}
        Quantitative evaluation of harmonisation quality using structural similarity index (SSIM$\uparrow$) and peak signal-to-noise ratio (PSNR$\uparrow$), measured against scanner-matched reference images.
        \textbf{c,}
        Downstream evaluation on the same cohort using brain-age prediction (MAE$\downarrow$) and scanner/site classification (Acc$\downarrow$). Effective harmonisation reduces scanner-specific information, reflected by lower site-classification accuracy, while preserving biologically relevant features required for accurate brain-age estimation.
        }
    \label{fig:cross_scanner}
\end{figure}

DIST-CLIP/I and DIST-CLIP/T achieved the highest SSIM and PSNR ($p<0.001$ vs. all baselines), indicating superior reconstruction quality under cross-site translation. On downstream tasks, unharmonised data yielded a brain-age MAE of 8.80 years and site classification accuracy of 0.79. DIST-CLIP/I reduced site classification accuracy to 0.42 while preserving brain-age performance (MAE 8.89 years). TUMSyn showed weaker scanner fingerprint removal (accuracy 0.45) and slightly higher MAE (8.96 years), consistent with its lower SSIM/PSNR and contrast discrepancies (Fig.~\ref{fig:cross_scanner}). HACA3 retained substantial site-specific signal (accuracy 0.78) and produced the highest brain-age error (9.13 years).

\begin{figure}[h]
    \centering
    \includegraphics[width=\linewidth]{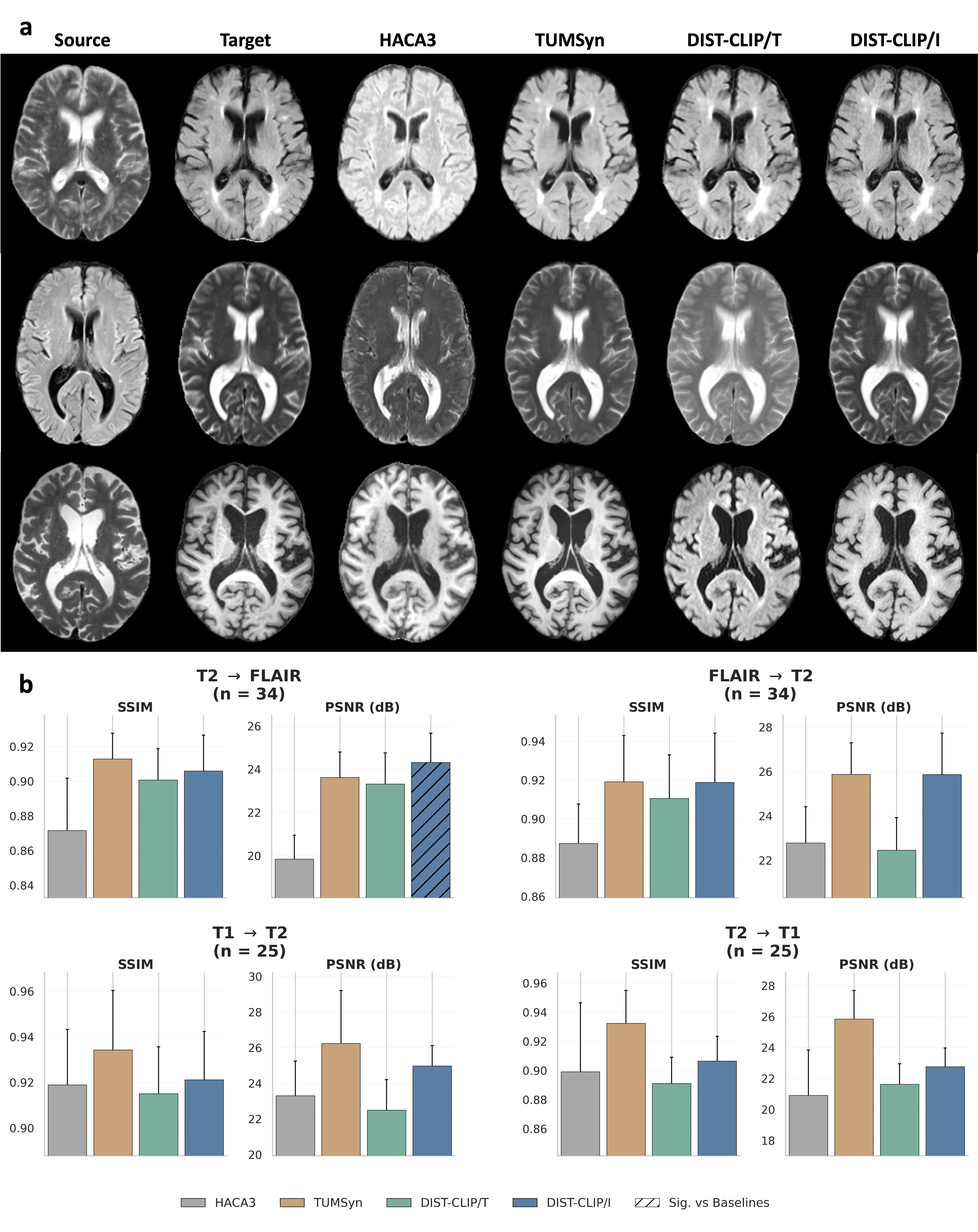}
    \caption{
    \textbf{Zero-shot harmonisation on OASIS-3.}
    \textbf{a,}
    Representative FLAIR$\rightarrow$T2w, T2w$\rightarrow$FLAIR and T1w$\rightarrow$T2w examples (source, target, HACA3, TUMSyn, DIST-CLIP/T and DIST-CLIP/I).
    \textbf{b,}
    Mean SSIM and PSNR ($\pm$s.d.; $n$ per panel) for the same translation directions. Hatched results indicate statistically significant improvements over the baselines ($p < 0.05$). n correspond to number of one-sided paired translations.
    }
    \label{fig:supp_oasis}
\end{figure}

Supplementary Fig. \ref{fig:supp_oasis} evaluates zero-shot harmonisation on OASIS-3, which was excluded from DIST-CLIP training; conversely, HACA3 and TUMSyn were trained directly on this cohort. Despite this distribution shift, DIST-CLIP/T and DIST-CLIP/I achieve SSIM and PSNR comparable to, and in some aspects exceeding, the in-distribution baselines across the shown translation directions (FLAIR$\leftrightarrow$T2w and T1w$\rightarrow$T2w). DIST-CLIP/I significantly outperforms HACA3 on both T1w$\rightarrow$T2w and FLAIR$\rightarrow$T2w tasks ($p < 0.001$ for SSIM and PSNR). On the FLAIR$\rightarrow$T2w translation, the zero-shot DIST-CLIP/I achieves performance that is statistically indistinguishable from the in-domain TUMSyn baseline (PSNR $p>0.05$, SSIM $p>0.05$).  Combined with its strong segmentation agreement (Fig. \ref{fig:cross_contrast}c), these findings support robust generalisation beyond the training distribution.

\subsection{Qualitative examples of harmonisation with pathological cases}

\begin{figure}[p]
    \centering
    \includegraphics[width=0.99\textwidth]{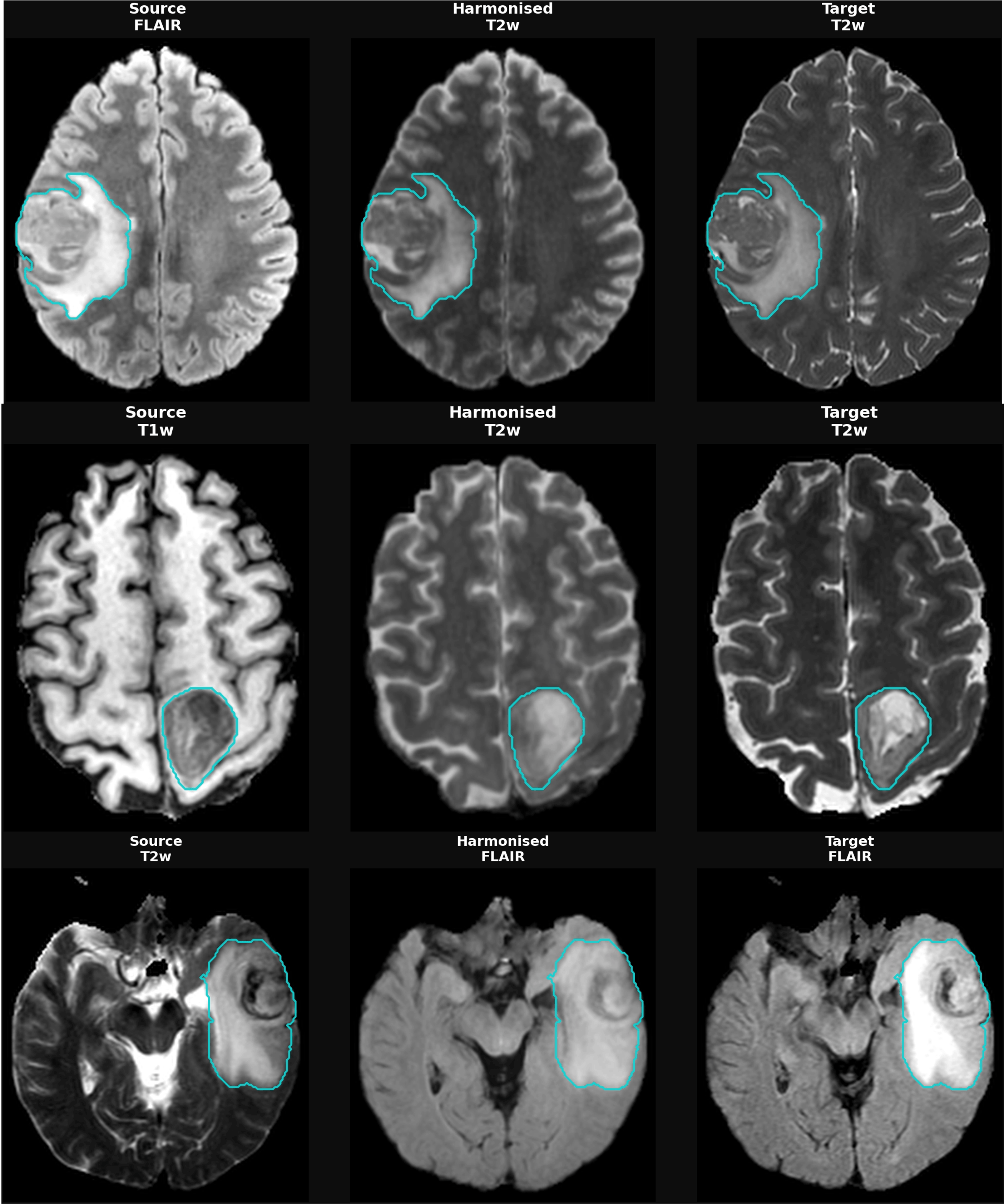}
\caption{\textbf{Tumour preservation under cross-contrast harmonisation.} Representative examples of cross-contrast harmonisation in brain tumour cases from the BraTS 2021 dataset.}
    \label{fig:supp_brats}
\end{figure}

We assessed pathology preservation on independent brain tumour cases from the BraTS 2021 dataset~\cite{baid2021rsna}. Supp. Fig.~\ref{fig:supp_brats} shows representative harmonisation results for three cross-contrast translations: $\text{FLAIR} \rightarrow \text{T2w}$, $\text{T1w} \rightarrow \text{T2w}$ and $\text{T2w} \rightarrow \text{FLAIR}$. These cases provide a challenging out-of-distribution evaluation because gliomas exhibit heterogeneous tumour regions, necrosis, mass effect and extensive peritumoral oedema. Despite residual imperfections in some translations, the tumour regions and surrounding pathological structures remain visible after harmonisation, including in cases with substantial differences between source and target contrasts.

\begin{figure}[p]
\centering
\includegraphics[width=\linewidth]{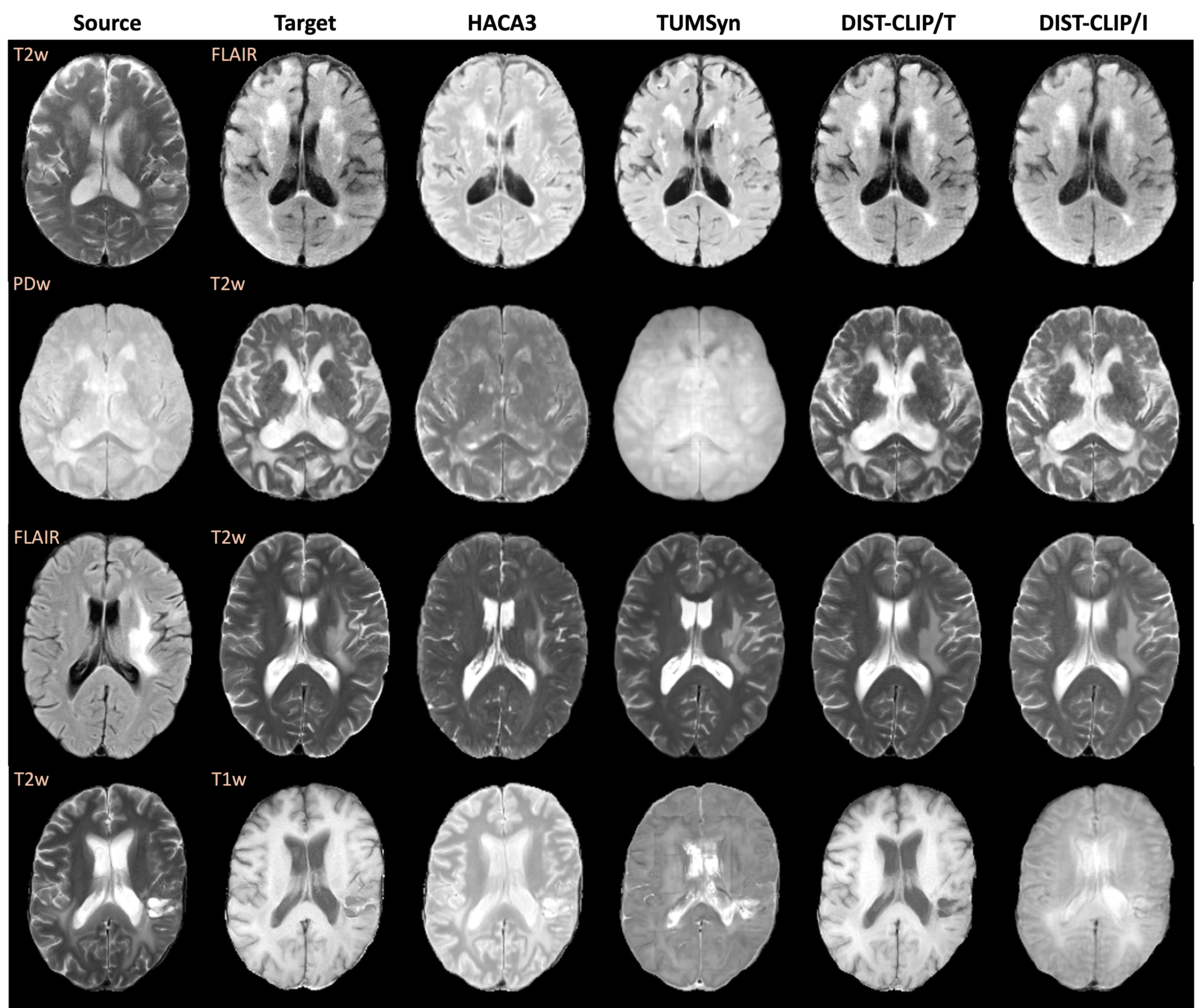}
\caption{\textbf{Pathology preservation on held-out cases.} Representative held-out cases comparing raw and harmonised images across cross-contrast translations.}
\label{fig:pat}
\end{figure}

We also evaluated pathology preservation on held-out test cases via qualitative comparison against the corresponding native acquisitions (Supp. Fig.~\ref{fig:pat}). While baseline methods frequently exhibit noticeable contrast discrepancies, DIST-CLIP achieves more faithful target-contrast synthesis while reliably preserving underlying anatomical and pathological structures. Nevertheless, residual failure modes persist, most notably in challenging regimens such as $\text{T2w} \rightarrow \text{T1w}$ translation, highlighting the inherent difficulty of cross-contrast harmonisation across substantial domain and appearance shifts.

\subsection{Style fusion decoder architecture}

\begin{figure}[p]
    \centering
    \includegraphics[width=\linewidth]{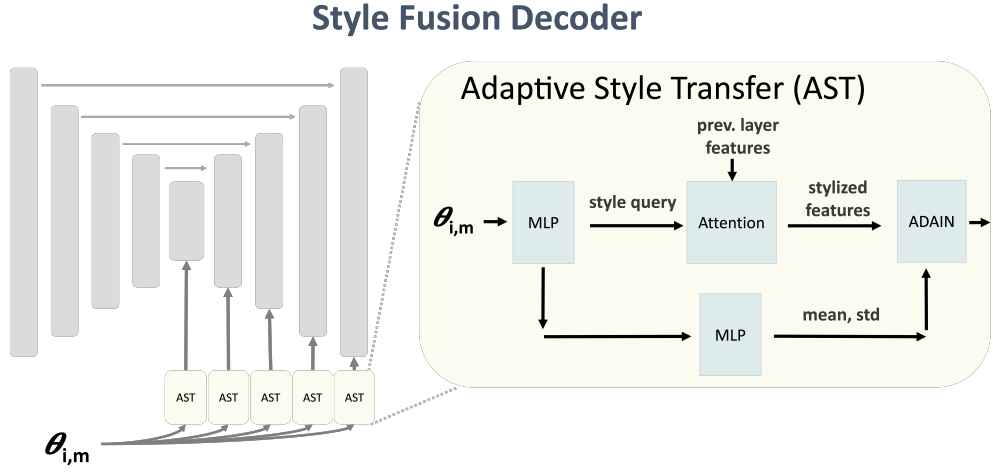}
    \caption{
    \textbf{Style fusion decoder and adaptive style transfer (AST).}
    \textbf{Left,}
    Style Fusion Decoder pathway: anatomical features from $\beta_{\mathrm{src}}$ are manipulated through successive decoder blocks with encoder skip connections; an AST block at each resolution injects the target contrast embedding $\theta$.
    \textbf{Right,}
    AST detail: $\theta$ is mapped by an MLP to a style query for attention over previous-layer features; a parallel MLP predicts AdaIN mean and standard deviation parameters that modulate the attended features to match target contrast appearance.
    }
    \label{fig:supp_sfd}
\end{figure}

Supplementary Fig.~\ref{fig:supp_sfd} summarises the multi-scale Style Fusion Decoder (SFD) and the internal AST module used for contrast injection. The SFD is built upon a 4-level U-Net backbone with base channel width $C_{\text{base}}=16$ (doubling at each level up to 512 at the bottleneck) integrated with AST modules across all five operational scales. Feature and cross-attention embedding dimensions match the decoder channel widths across scales: 512 at the bottleneck ($14\times 14$), followed by 256, 128, 64, and 32 along the upsampling path ($28\times 28$ to $224\times 224$). Each scale employs 8 attention heads, yielding head dimensions of 64, 32, 16, 8, and 4, respectively, using multi-head attention at the bottleneck and linear windowed attention (window size 16) at higher resolutions. In each AST block, a two-layer MLP projects the 512-dimensional CLIP style embedding into channel dimension $C$ via an intermediate expansion ($512 \to 2C \to C$). The style and image features are concatenated ($2C$ channels) and passed through a dual-convolution head ($\text{Conv}{2C \to C} \to \text{ReLU} \to \text{Conv}{C \to 2C}$) to predict $C$-channel spatial modulation parameters $(\gamma, \beta)$.

\subsection{Held-out harmonisation test pairs}

Supplementary Table~\ref{tab:harm_test_pairs} lists the number of source--target contrast pairs in the held-out harmonisation test set for each translation direction (from the DIST-CLIP evaluation protocol).

\begin{table}[h]
\centering
\caption{Held-out test source--target pair counts for cross-contrast harmonisation.}
\label{tab:harm_test_pairs}
\begin{tabular}{lcccc}
\toprule
Source $\rightarrow$ target & T1w & T2w & PDw & FLAIR \\
\midrule
T1w  & --   & 50  & 1   & 26 \\
T2w  & 50  & --  & 274 & 366 \\
PDw  & 1   & 274 & --  & 41 \\
FLAIR & 26  & 366 & 41  & -- \\
\bottomrule
\end{tabular}
\end{table}

\subsection{Alignment of synthetic images with physical expectations}

We examined whether synthetic volumes generated under different metadata conditions retain the corresponding contrast and scanner characteristics in a pretrained MR-CLIP feature space. Such structure would indicate that metadata conditioning is reflected in the resulting images and that the generator can independently modulate sequence contrast and acquisition-related characteristics.

\begin{figure}[p]
\centering
\includegraphics[width=\linewidth]{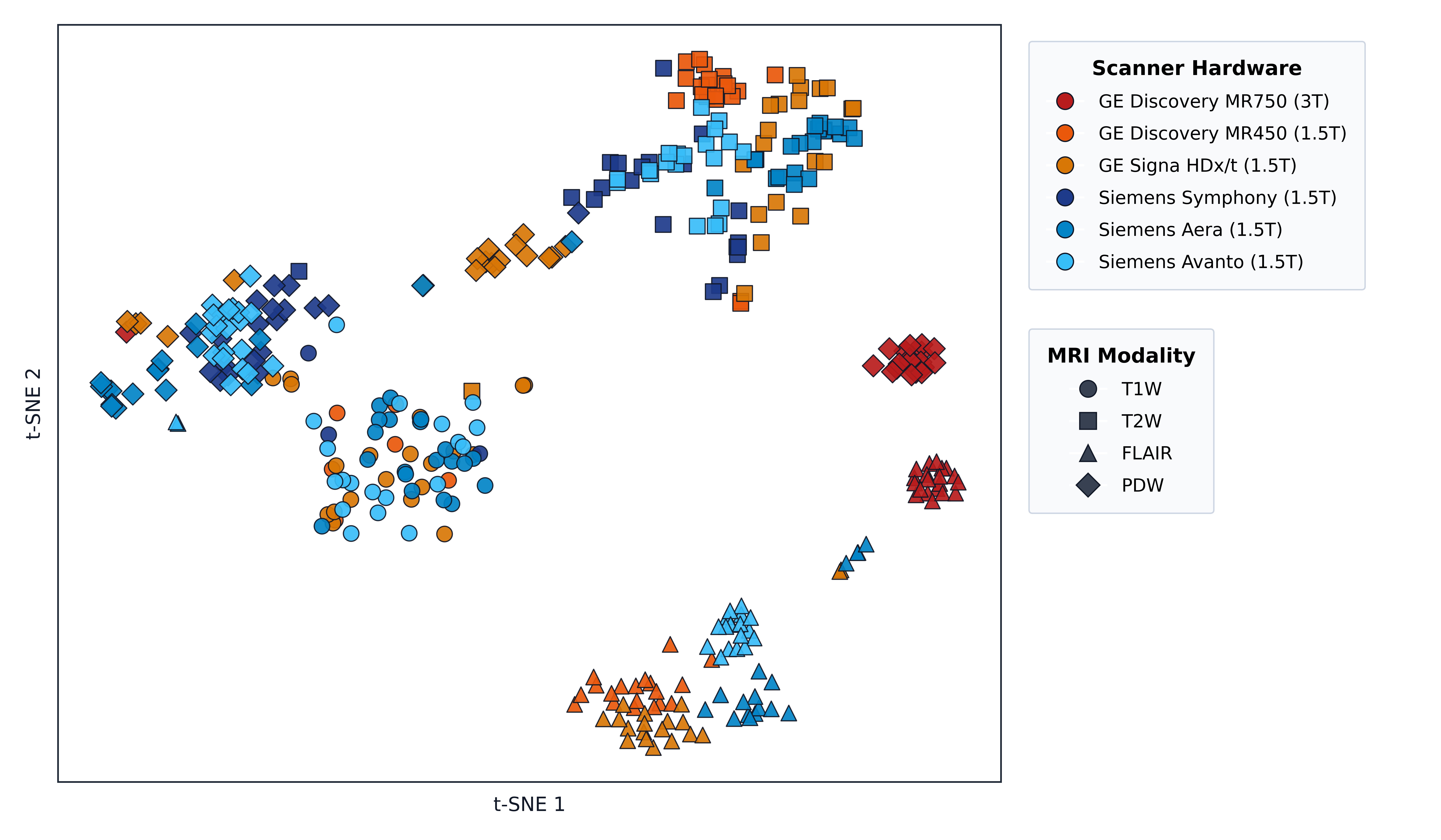}
\caption{\textbf{Feature-space structure of metadata-conditioned synthetic volumes.}
t-SNE projection of MR-CLIP embeddings extracted from synthetic volumes generated across four contrast weightings (T1w, T2w, FLAIR and PDw) and six scanner platforms. Each point represents a synthetic volume. Contrast weightings form the dominant structure in the embedding space, with additional organisation by scanner platform and field strength within contrast groups.}
\label{fig:scanner_tsne}
\end{figure}

\begin{figure}[p]
\centering
\includegraphics[width=\linewidth]{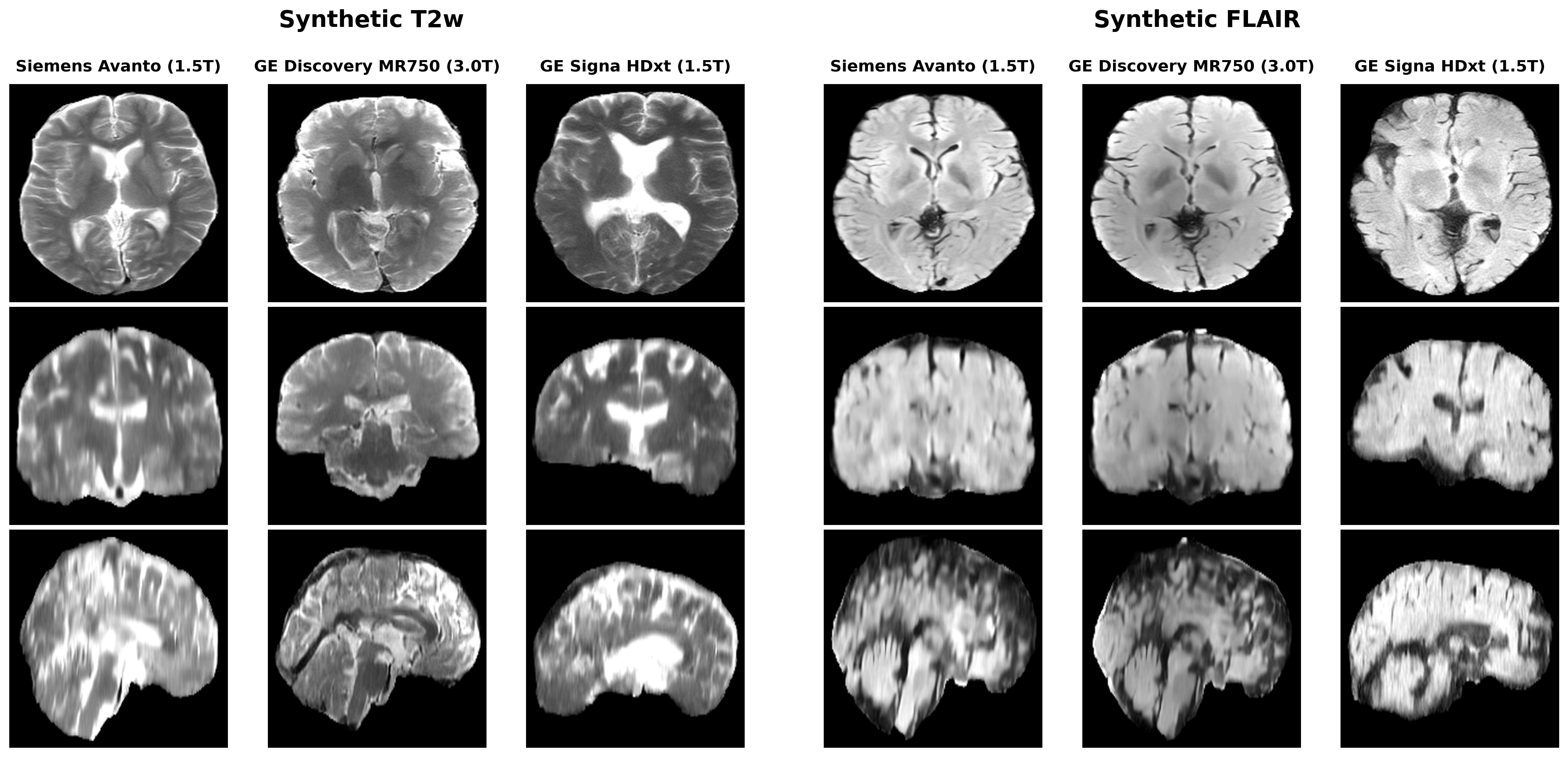}
\caption{\textbf{Metadata-conditioned synthesis across scanners.}
Exemplary slices from synthetic volumes generated from the corresponding acquisition metadata. Examples span T2w, PDw, and FLAIR acquisitions across Siemens Avanto (1.5T), GE Signa HDx (1.5T), and Siemens GE Discovery (3T). The synthetic volumes reproduce characteristic contrast differences across sequences, including bright CSF in T2w,, and CSF suppression in FLAIR, while retaining realistic brain anatomy across scanner conditions.}
\label{fig:real_vs_synth_matrix}
\end{figure}

The embeddings primarily separated according to contrast weighting, with T1w, T2w, FLAIR and PDw forming distinct regions of the manifold (Fig.~\ref{fig:scanner_tsne}). Within these regions, samples further organised according to scanner platform and field strength, including separation between 1.5T and 3T systems and clustering of scanners from the same manufacturer. Thus, the generated images preserve both sequence-dependent contrast and scanner-associated variation specified by the conditioning metadata. Further visual examples are provided in Fig. \ref{fig:real_vs_synth_matrix}.

\begin{figure}[p] \centering \includegraphics[width=\linewidth]{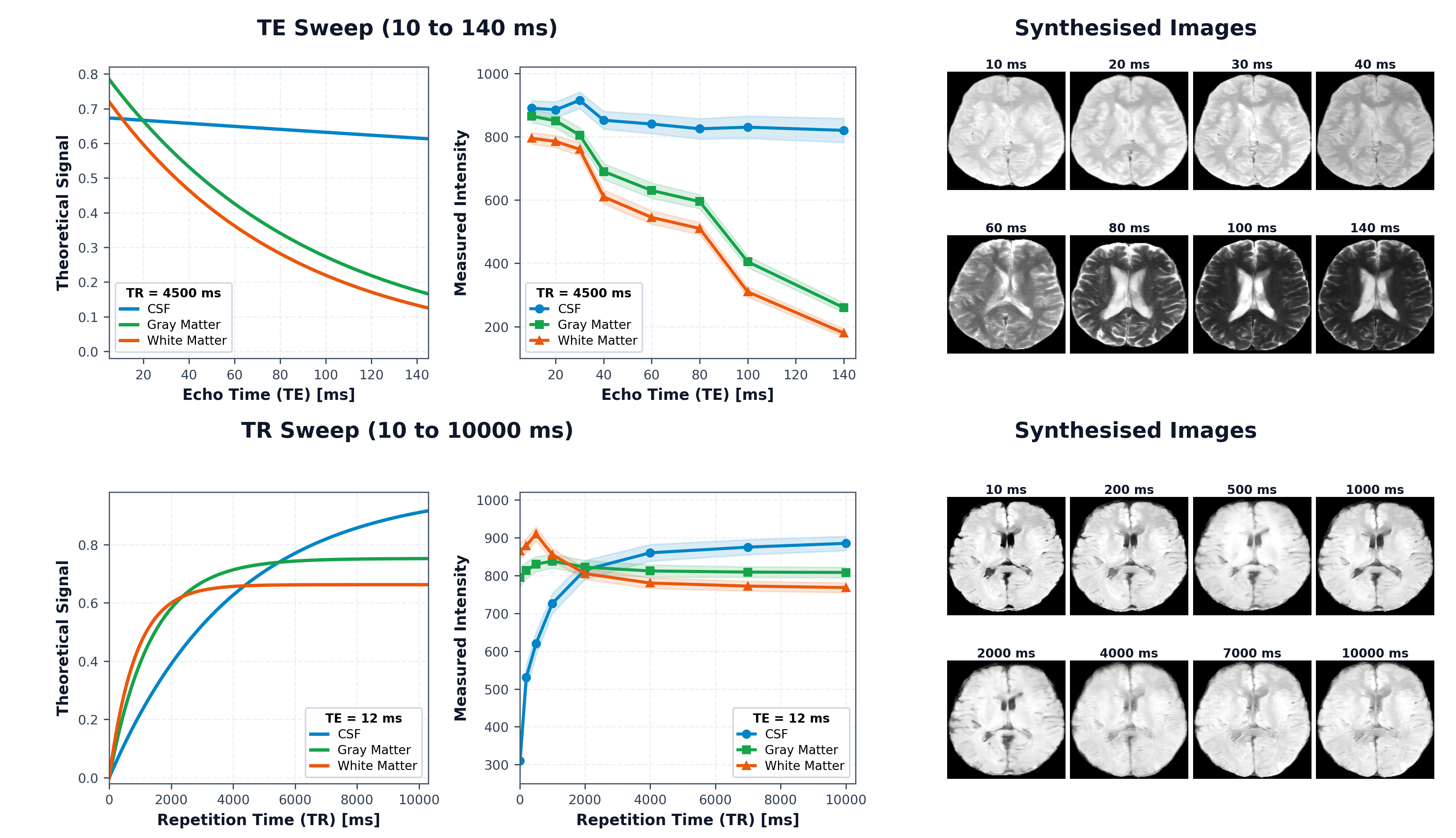} 
\caption{ 
\textbf{Biophysical consistency of metadata-conditioned synthesis under continuous TE and TR sweeps.}  Measured tissue intensity curves compared to theoretical expectations along with axial slices \textbf{(a)} generated by continuously varying $\mathrm{TE}$ from $10$ to $140\,\mathrm{ms}$ while keeping $\mathrm{TR}=4500\,\mathrm{ms}$ fixed and using the same anatomical latent code, 
\textbf{(b)} generated by varying $\mathrm{TR}$ from $10$ to $10000\,\mathrm{ms}$ while keeping $\mathrm{TE}=12\,\mathrm{ms}$ fixed. 
} 
\label{fig:physics_te_tr_sweeps} 
\end{figure}

Discrete sequence labels alone do not establish whether the model has learned a meaningful relationship between acquisition parameters and image appearance. We therefore varied numerical parameters continuously while keeping the remaining conditioning and latent noise fixed, and assessed whether the resulting changes in image contrast followed expected MR signal behaviour (Supp. Fig.~\ref{fig:physics_te_tr_sweeps}). For each conditioning, we generated five images and report their mean tissue intensities. Sweeping $\text{TE}$ from $10$--$140 \text{ms}$ produced progressively greater signal attenuation, particularly in white and grey matter, consistent with their shorter effective $\text{T}_2$ values relative to CSF. Increasing $\text{TR}$ from $10$ to $10{,}000 \text{ms}$ progressively reduced longitudinal saturation and altered relative tissue contrast, consistent with increasing longitudinal recovery. The absolute WM and GM intensity response for TR swep does not directly follow the theoretical signal magnitude. Despite the near-zero longitudinal magnetisation expected at very short TR, WM and GM retained relatively high intensities in the generated images. This may reflect the intensity scaling present in clinical MRI data, potentially including prescan receiver-gain adjustment, which can rescale signal amplitudes before digitisation. As the model is trained on clinical DICOM images, such intensity characteristics are part of the learned image distribution. Importantly, the relative contrast remains acquisition-dependent: CSF is strongly suppressed at short TR ($10$--$200 \text{ms}$) and progressively recovers with increasing TR, becoming hyperintense at long TR ($>2{,}000 \text{ms}$), consistent with its long $T_1$ relaxation time.

\subsection{Multimodal latent space alignment across acquisition parameters}
\label{sec:tsne_analysis}

\begin{figure}[p]
\centering
\includegraphics[width=\textwidth]{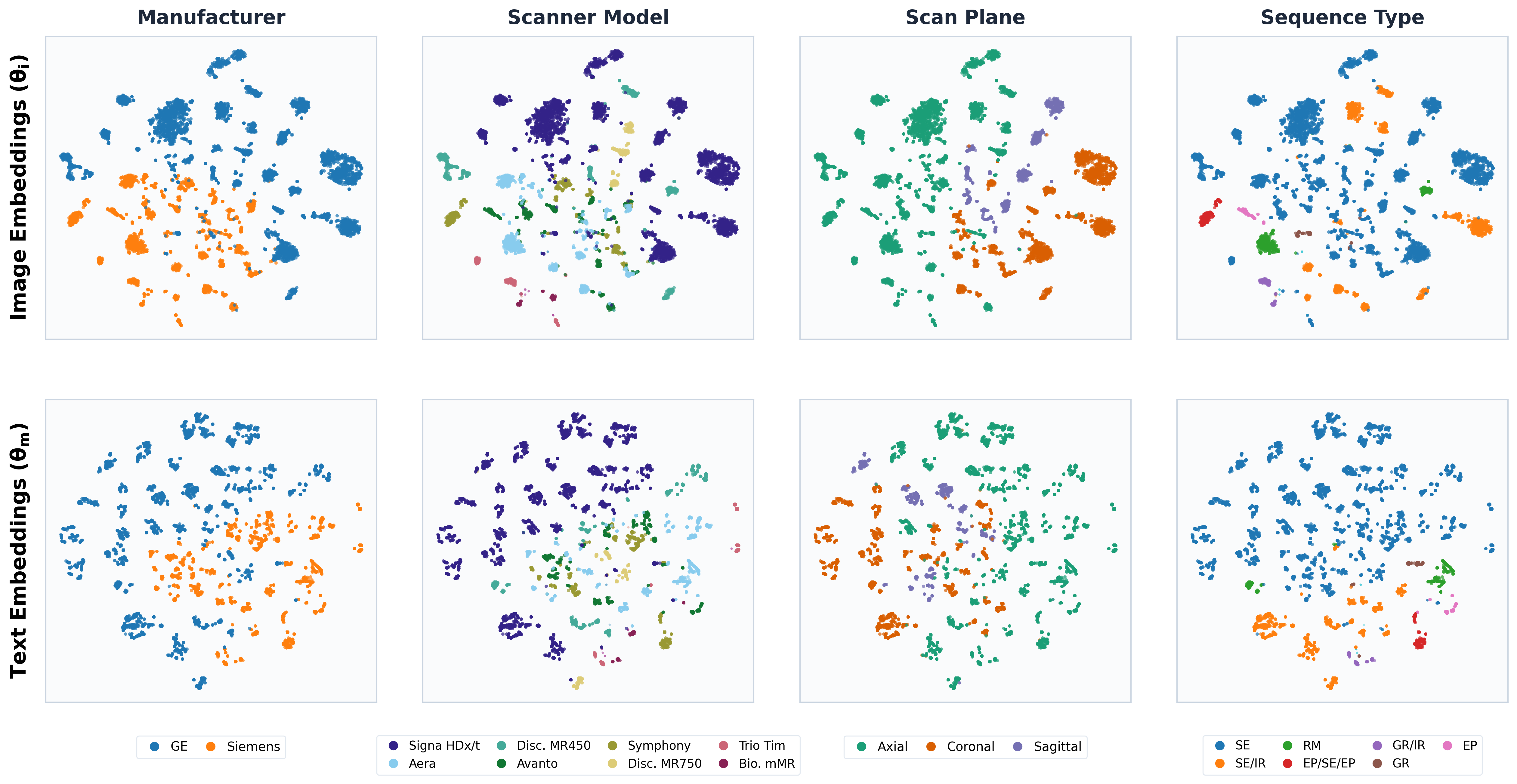}
\caption{\textbf{Multimodal latent space representation geometry across acquisition specifications.} 
Two-dimensional t-SNE projections of held-out test set embeddings. The top row visualises image contrast representations ($\mathbf{\theta}_i$), and the bottom row illustrates metadata representations  ($\mathbf{\theta}_m$). Columns depict distributions stratified across four key DICOM acquisition specifications: scanner manufacturer, scanner model, scan plane, and sequence type. Both modalities exhibit closely mirrored, structured manifold geometry that preserves physical acquisition properties without explicit categorical supervision.}
\label{fig:tsne_multimodal}
\end{figure}

To investigate whether MR-CLIP retains coherent physical and protocol-level semantics within its learned multimodal embedding space, we visualised two-dimensional $t$-SNE projections of held-out test embeddings across both visual and textual modalities as illustrated in Fig. ~\ref{fig:tsne_multimodal}. Both embedding manifolds demonstrate distinct, physically consistent topological clustering. At the macro level, the representation spaces naturally separate by scanner vendor (GE vs.\ Siemens) and spatial orientation (axial, coronal, and sagittal planes). At a finer granularity, specific scanner hardware families (e.g., Signa HDx/t, Aera, Avanto) and sequence physics categories (e.g., Spin Echo (SE), Inversion Recovery (SE/IR), Gradient Echo (GR) form coherent sub-clusters. Structural clustering evident in the image embedding space ($\mathbf{\theta}_i$) closely parallels the geometry observed in the metadata embedding space ($\mathbf{\theta}_m$), confirming that MR-CLIP successfully aligns rich textual and numerical acquisition metadata with volumetric MRI features in latent space.

\subsection{Ablation studies}
Main experiments used SynthStrip skull-stripped brain volumes. To quantify sensitivity to this preprocessing choice, we retrained and re-evaluated models on matched data with the skull retained, keeping registration, resampling, train/test splits and evaluation protocols otherwise identical.

We compared three-dimensional MR-CLIP on the held-out test set under skull-stripped versus with-skull (Supplementary Table~\ref{tab:skull_ablation_retrieval}). Retaining the skull improved retrieval scores which shows that non-brain tissues such as the skull carry appearance variations that are directly modulated by acquisition parameters: volume-to-metadata R@1 increased slightly (62.2\% versus 60.2\%), along with metadata-to-volume matching (R@1 87.3\% versus 79.3\%; R@5 96.7\% versus 91.6\%).

\begin{table}[h]
\centering
\small
\caption{Skull-stripping ablation for three-dimensional MR-CLIP cross-modal retrieval on the KCH held-out test set.
Recall at ranks 1, 5 and 10 (R@1/5/10, \%) for volume-to-metadata (V$\rightarrow$M) and metadata-to-volume (M$\rightarrow$V) matching. Models were trained and evaluated under the same preprocessing setting in each row.}
\label{tab:skull_ablation_retrieval}
\begin{tabular}{@{}lcccccc@{}}
\toprule
 & \multicolumn{3}{c}{V$\rightarrow$M} & \multicolumn{3}{c}{M$\rightarrow$V} \\
\cmidrule(lr){2-4}\cmidrule(lr){5-7}
Setting & R@1 & R@5 & R@10 & R@1 & R@5 & R@10 \\
\midrule
Skull-stripped & 60.2 & 79.0 & 82.0 & 79.3 & 91.6 & 94.0 \\
With skull     & 62.2 & 78.8 & 82.7 & 87.3 & 96.7 & 97.7 \\
\bottomrule
\end{tabular}
\end{table}

We repeated held-out cross-contrast harmonisation with DIST-CLIP/I under the same pair set (Supplementary Table~\ref{tab:skull_ablation_ssim_psnr}). Skull-stripping markedly improved structural agreement (mean SSIM $0.952 \pm 0.020$ versus $0.828 \pm 0.057$ with skull retained), whereas mean PSNR was similar ($27.91 \pm 3.96$\,dB versus $27.64 \pm 2.55$\,dB). The larger SSIM gain indicates that non-brain intensity structure in with-skull inputs mainly harms spatial correspondence to the target contrast, not global intensity scale.

\begin{table}[h]
\centering
\caption{Skull-stripping ablation on held-out harmonisation (DIST-CLIP/I).
Mean $\pm$ s.d.\ over harmonisation pairs present in both the skull-stripped and with-skull.}
\label{tab:skull_ablation_ssim_psnr}
\begin{tabular}{lcc}
\toprule
Setting & SSIM $\uparrow$ & PSNR (dB) $\uparrow$ \\
\midrule
Skull-stripped & $0.952 \pm 0.020$ & $27.91 \pm 3.96$ \\
With skull     & $0.828 \pm 0.057$ & $27.64 \pm 2.55$ \\
\bottomrule
\end{tabular}
\end{table}

We additionally evaluated the sensitivity of MR-CLIP to the choice of text encoder architecture under the 3D setting trained with skull and demographic metadata (Table~\ref{tab:text_encoder_ablation_retrieval}). The default text transformer and TARTE achieved broadly similar scan-to-metadata retrieval performance, with TARTE yielding a slightly higher top-1 recall (62.8\% vs.\ 62.2\%), while the transformer achieved higher top-5 and top-10 recall (78.8\% and 82.7\% vs.\ 77.4\% and 82.4\%). For metadata-to-scan retrieval, the transformer performed substantially better at top-1 (87.3\% vs.\ 80.9\%), while TARTE achieved comparable top-5 and top-10 recall (96.5\% and 97.9\% vs.\ 96.7\% and 97.7\%). In contrast, the MLP encoder consistently underperformed both architectures, particularly for metadata-to-scan retrieval, where top-1 recall decreased to 47.2\%. These results indicate that the choice of text encoder has limited impact when using the transformer or TARTE, whereas the simpler MLP provides substantially weaker cross-modal alignment.

\begin{table}[h]
\centering
\small
\caption{Comparison of different text encoder architectures for three-dimensional MR-CLIP cross-modal retrieval on the held-out test set (trained with skull and age/sex metadata).}
\label{tab:text_encoder_ablation_retrieval}
\begin{tabular}{@{}lcccccc@{}}
\toprule
 & \multicolumn{3}{c}{Scan$\rightarrow$Metadata} & \multicolumn{3}{c}{Metadata$\rightarrow$Scan} \\
\cmidrule(lr){2-4}\cmidrule(lr){5-7}
Text Encoder & R@1 & R@5 & R@10 & R@1 & R@5 & R@10 \\
\midrule
Transformer & 62.2 & \textbf{78.8} & \textbf{82.7} & \textbf{87.3} & \textbf{96.7} & 97.7 \\
MLP                     & 55.3 & 72.9 & 78.6 & 47.2 & 70.6 & 76.6 \\
TARTE                        & \textbf{62.8} & 77.4 & 82.4 & 80.9 & 96.5 & \textbf{97.9} \\
\bottomrule
\end{tabular}
\end{table}

\begin{figure}
    \centering
    \includegraphics[width=0.99\columnwidth]{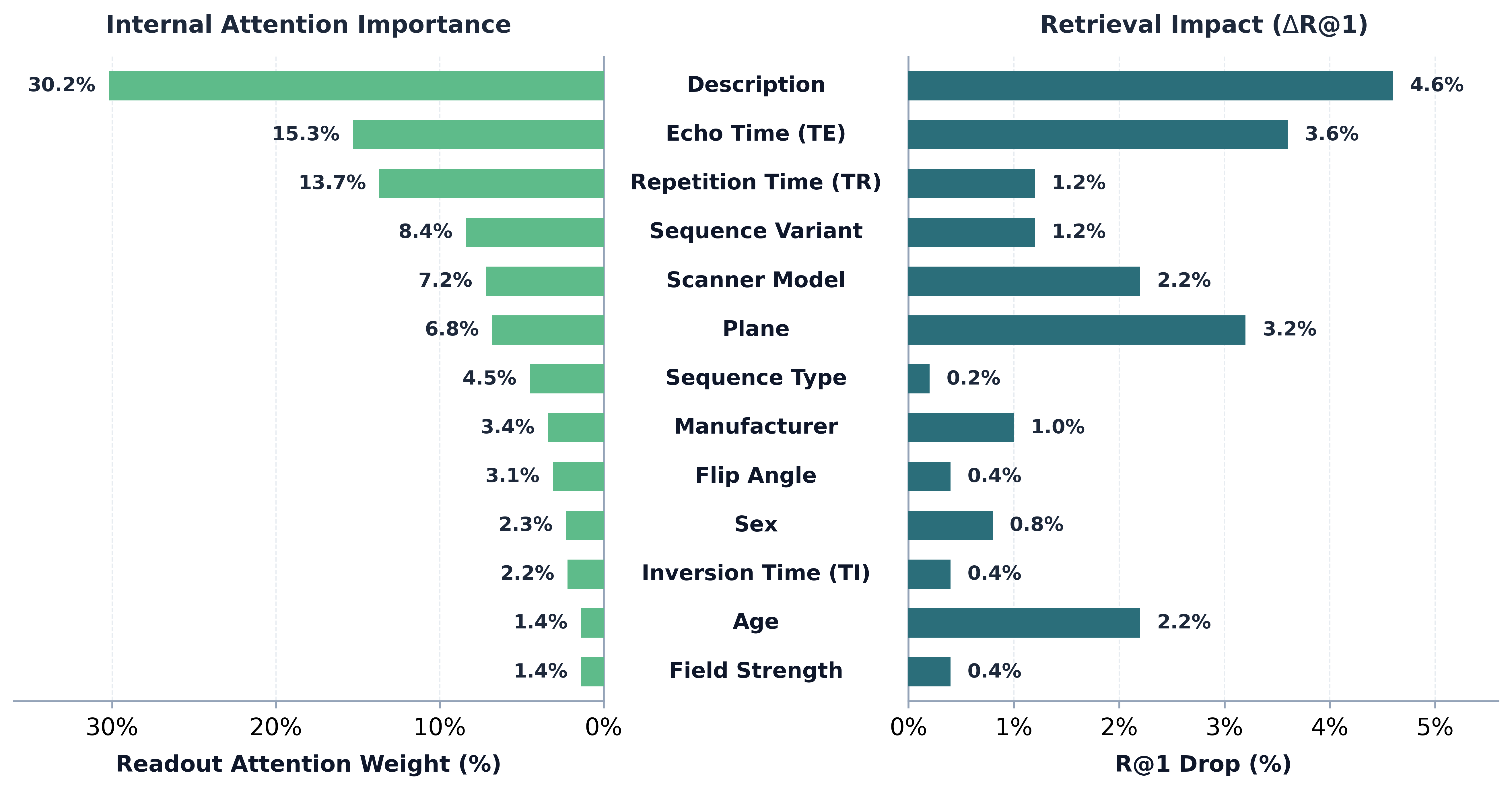}
\caption{\textbf{Metadata field importance for the TARTE encoder.} \textbf{Left:} relative internal attention weight assigned by the TARTE text encoder to each metadata field, averaged over the readout attention across the test set. \textbf{Right:} corresponding drop in Scan$\rightarrow$Metadata Recall@1 (R@1) when each field is individually left out at test time, quantifying its marginal contribution to retrieval performance.}

    \label{fig:supp_att}
\end{figure}

To understand how the TARTE module aggregates multimodal tabular metadata, we analysed both the self-attention weights assigned by the aggregation readout token in the final transformer layer and the drop in Scan
→Metadata Recall@1 when each field is individually dropped at test time (Fig.~\ref{fig:supp_att}). The network relies most heavily on the Series Description (30.2\% attention; 4.6-point R@1 drop), which encapsulates global protocol semantics, followed by the core physical acquisition parameters governing tissue contrast, Echo Time (TE, 15.3\%; 3.6 points) and Repetition Time (TR, 13.7\%; 1.2 points). Although Series Description is individually the most informative tag, the acquisition physics parameters, Echo Time ($\text{TE}$), Repetition Time ($\text{TR}$), Inversion Time ($\text{TI}$), Flip Angle, collectively play a more dominant role. This joint reliance directly reflects MRI physics: tissue contrast ($\text{T}_1$, $\text{T}_2$, FLAIR) is not dictated by any single parameter in isolation, but emerges from the coupled interaction of $\text{TE}$, $\text{TR}$, $\text{TI}$, and flip angle within the pulse sequence. Additional secondary context also comes from structural and scanner-specific tags. Non-contrast demographics receive minimal attention overall, though Age is a notable exception: despite its low attention weight (1.4\%), dropping it produces a disproportionately large retrieval drop (2.2 points). 

Furthermore, we investigate the contribution of DIST-CLIP components through an ablation study summarised in Supp. Table \ref{tab:ablation}. Replacing the AST blocks with a standard AdaIN layer results in a significant performance drop, indicating that attention-driven modulation is crucial for precise contrast transfer. We also evaluate the setting without anatomical disentanglement, where the source image is directly fed into the SFD without passing through the Anatomy Mapper which leads to small reduction in PSNR and SSIM. We further analyse the contribution of different guidance modalities. In the only metadata guidance and only image guidance settings, the corresponding modality is not provided as an explicit style input to the decoder, but it is still incorporated through the CLIP-based alignment loss during training. As a result, the model can still benefit from weak supervisory signals from the unused modality, which explains the relatively small performance differences across these variants. Overall, the full DIST-CLIP model, which jointly leverages both image-based and metadata-based guidance at the architectural level and through CLIP supervision, achieves the most balanced and consistent performance across PSNR and SSIM, demonstrating that complementary multi-modal guidance improves robust harmonisation.

\begin{table}[h]
\centering
\caption{Ablation study of DIST-CLIP components showing image and text guided performance, averaged over all contrast translation tasks.}
\label{tab:ablation}
\begin{tabular}{lcccc}
\toprule
& \multicolumn{2}{c}{\textbf{PSNR} ↑} & \multicolumn{2}{c}{\textbf{SSIM} ↑} \\
\cmidrule(lr){2-3} \cmidrule(lr){4-5}
\textbf{Model Variant} 
& \textbf{Image} & \textbf{Text} 
& \textbf{Image} & \textbf{Text} \\
\midrule
AdaIN instead of AST blocks                   & 21.6 & 21.9 & 0.889 & 0.900 \\
Without Anatomy Mapper & 28.5 & 28.2 & 0.949 & 0.950 \\
Trained with only metadata            & 27.6 & 28.3 & 0.941 & 0.950 \\
Trained with only image               & \textbf{29.0} & 27.7 & \textbf{0.953} & 0.949 \\
\textbf{Full DIST-CLIP}                        & 28.7 & \textbf{28.4} & 0.951 & \textbf{0.951} \\
\bottomrule
\end{tabular}
\end{table}

\subsection{Grouping strategy of numerical tags}
 We investigate how discretising the TE–TR–TI space affects retrieval performance (Supplementary Table \ref{tab:grid_transfer}). We evaluate both in-domain settings (same grid for training and testing) and cross-discretisation generalisation. Overall, coarser groupings consistently improve performance. In particular, a 5×5 TE–TR grid achieves the best results across most metrics, which indicates that reduced label granularity simplifies the alignment task, likely by lowering label noise and increasing intra-class consistency. However, lowering grid size results in loss of fine-grained understanding of the parameter space; therefore 20x20 was chosen in rest of the experiments. We further compare grid-based discretisation with k-means clustering. Although k-means with 20 clusters outperforms the closest grid setting (5×5) for image-to-text and 3D scan-to-text retrieval, it performs worse on text-to-image retrieval. Increasing the number of clusters also leads to slightly degraded performance relative to comparable grid resolutions. These results suggest that k-means is sensitive to the choice of cluster number and tightly coupled to the training distribution, limiting its robustness across acquisition protocols. In contrast, grid-based grouping, while not always optimal, is more stable and likely to generalise better to diverse clinical settings.

Finally, we assess transferability by training MR-CLIP on a fine 20×20 grid and evaluating it on coarser discretisations without retraining. Despite being trained on fine-grained labels, the model maintains strong performance on coarser groupings, demonstrating robustness and flexibility across different settings.

\begin{table*}[t]
\centering
\caption{Retrieval performance (\%) across different discretisation granularities of TE and TR (TExTR). Top: results when training and evaluating on the same set. Bottom: generalisation performance when transferring from a $20\times20$ training grid to other discretisations. We report R@1/5/10 for slice-to-metadata, 3D scan-to-metadata, and metadata-to-slice retrieval. Highest values in each column are bolded.}
\resizebox{\textwidth}{!}{
\begin{tabular}{l @{\hspace{0.4em}} ccc @{\hspace{0.4em}}|@{\hspace{0.4em}} ccc @{\hspace{0.4em}}|@{\hspace{0.4em}}ccc}
\toprule
\textbf{Set (\# total classes)} & \multicolumn{3}{c}{\textbf{Slice$\rightarrow$Metadata}} & \multicolumn{3}{c}{\textbf{3D Scan$\rightarrow$Metadata}} & \multicolumn{3}{c}{\textbf{Metadata$\rightarrow$Slice}} \\
\cmidrule(lr){2-4} \cmidrule(lr){5-7} \cmidrule(lr){8-10}
 & \multicolumn{1}{c}{R@1} & \multicolumn{1}{c}{R@5} & \multicolumn{1}{c}{R@10} & \multicolumn{1}{c}{R@1} & \multicolumn{1}{c}{R@5} & \multicolumn{1}{c}{R@10} & \multicolumn{1}{c}{R@1} & \multicolumn{1}{c}{R@5} & \multicolumn{1}{c}{R@10} \\
\midrule
\multicolumn{10}{l}{\textit{Trained and evaluated on same sets}} \\
\midrule
40x20 (1770)    & 62.5 & 72.1 & 73.2 & 73.9 & 92.8 & 94.1 & 88.6 & 92.4 & 93.5 \\
20x20 (1415)    & 66.0 & 77.3 & 78.3 & 78.7 & 94.2 & 95.2 & 90.9 & 93.6 & 94.4 \\
20x10 (1017)    & 69.5 & 80.4 & 81.1 & 82.0 & 95.4 & 96.0 & 94.1 & 96.1 & 96.8 \\
10x10 (792)     & 77.2 & 84.3 & 85.2 & 86.4 & 96.1 & 96.6 & 92.0 & 96.5 & 97.6 \\
10x5 (599)      & 77.3 & 86.2 & 86.7 & 84.3 & 96.7 & 97.1 & 86.6 & 97.7 & 98.0 \\
5x5 (488)       & 69.7 & 86.9 & 88.0 & 89.1 & 97.0 & 97.4 & \textbf{93.7} & \textbf{97.8} & \textbf{98.1} \\
k-means(nc=100) (1220)  & 60.1 & 70.5 & 71.6 & 72.6 & 93.6 & 95.2 & 91.2 & 93.5 & 94.5 \\
k-means(nc=50) (826)  & 71.8 & 83.0 & 83.7 & 83.9 & 96.1 & 96.9 & 93.6 & 95.6 & 96.0 \\
k-means(nc=20) (522)  & \textbf{81.4} & \textbf{90.1} & \textbf{90.4} & \textbf{91.3} & \textbf{97.2} & \textbf{97.6} & 88.7 & 97.7 & 97.8 \\

\midrule
\multicolumn{10}{l}{\textit{Trained on 20x20 set, evaluated on other sets}} \\
\midrule
40x20 (1770)    & 54.1 & 74.1 & 76.7 & 68.0 & 93.2 & 94.7 & 74.1 & 84.4 & 88.4 \\
20x20 (1415)    & 66.0 & 77.3 & 78.3 & 78.7 & 94.2 & 95.2 & 90.9 & 93.6 & 94.4 \\
20x10 (1017)    & 71.1 & 80.1 & 80.8 & 81.7 & 95.4 & 96.1 & 93.2 & 95.2 & 95.7 \\
10x10 (792)     & 76.7 & 84.7 & 85.3 & 85.9 & 96.1 & 96.6 & 93.8 & 95.7 & 96.3 \\
10x5 (599)      & 78.9 & 86.2 & 86.6 & 87.3 & 96.6 & 97.1 & 94.7 & 96.5 & 97.0 \\
5x5 (488)       & \textbf{83.6} & \textbf{89.9} & \textbf{90.3} & \textbf{90.5} & \textbf{97.2} & \textbf{97.5} & \textbf{95.6} & \textbf{96.8} & \textbf{97.3} \\
\bottomrule
\end{tabular}
}
\label{tab:grid_transfer}
\end{table*}



\subsection{Quality-control detection performance}

\begin{figure}[h]
    \centering
    \includegraphics[width=\linewidth]{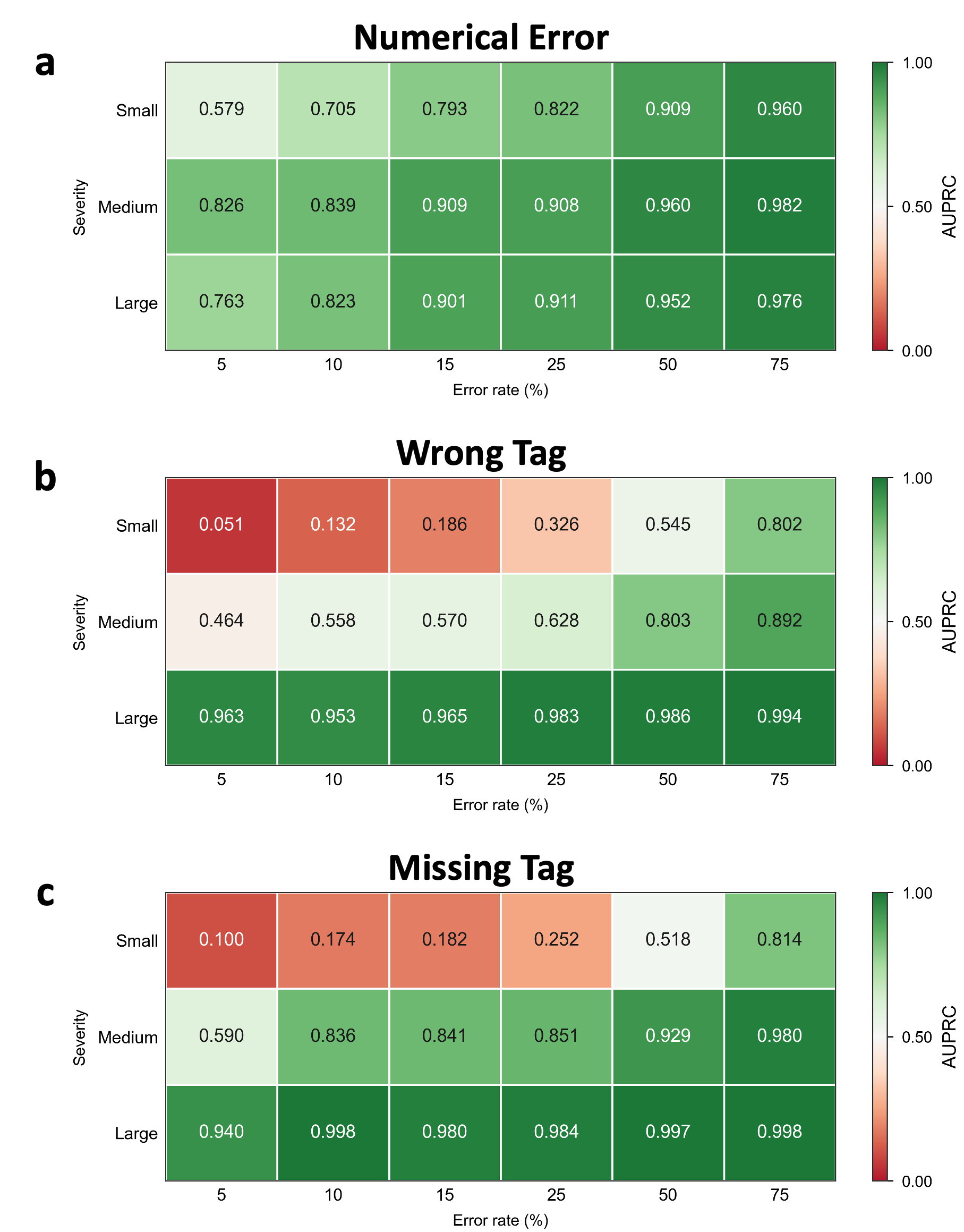}
    \caption{
    \textbf{Metadata corruption detection across corruption rates.}
    Detection AUC using one minus cosine similarity between MR-CLIP image and metadata embeddings as an anomaly score, evaluated on held-out test volumes with synthetic metadata corruptions at rates of 5--75\% (columns) and small, medium or large severity (rows).
    \textbf{a,}
    Numerical errors in echo time, repetition time and inversion time.
    \textbf{b,}
    Incorrect categorical tags (series description, sequence fields, scanner identifiers).
    \textbf{c,}
    Missing tags (progressive removal of the same fields).
    }
    \label{fig:supp_qc}
\end{figure}

Supplementary Fig.~\ref{fig:supp_qc} reports detection AUC (embedding distance as anomaly score) for numerical timing errors (\textbf{a}), incorrect categorical tags (\textbf{b}) and missing fields (\textbf{c}), evaluated at corruption rates of 5, 10, 15, 25, 50 and 75\% and at small, medium and large severity. Performance increased with both corruption rate and severity for all classes; large numerical and missing-tag edits remained highly detectable even at low prevalence, whereas subtle wrong-tag and missing-tag corruptions were hardest at low rates, consistent with the 50\% summary in Fig.~\ref{fig:qc}c. 

Notably, at small perturbation severities, performance is comparatively lower because minor synthetic corruptions alter the Series Description. This tag was deliberately excluded from the primary contrast group label construction and utilised solely as auxiliary conditioning metadata. In clinical practice, the Series Description is unstandardised free text that exhibits extreme variability across institutions, including site-specific naming conventions, non-standard abbreviations, and multi-lingual descriptors; an isolated textual discrepancy frequently reflects local documentation style rather than an authentic physical acquisition mismatch. Because supervisory contrast grouping is anchored strictly in physical pulse-sequence parameters (TE, TR, flip angle) rather than free-text sequence descriptors, the representation space is naturally less sensitive to isolated, fine-grained textual substitutions within the Series Description when underlying physical parameters remain unperturbed. Consequently, minor textual corruptions yield lower detection metrics, reflecting a deliberate design choice that prioritises physical protocol consistency over nominal label variations.

\begin{figure}[p]
\centering
\includegraphics[width=0.95\linewidth]{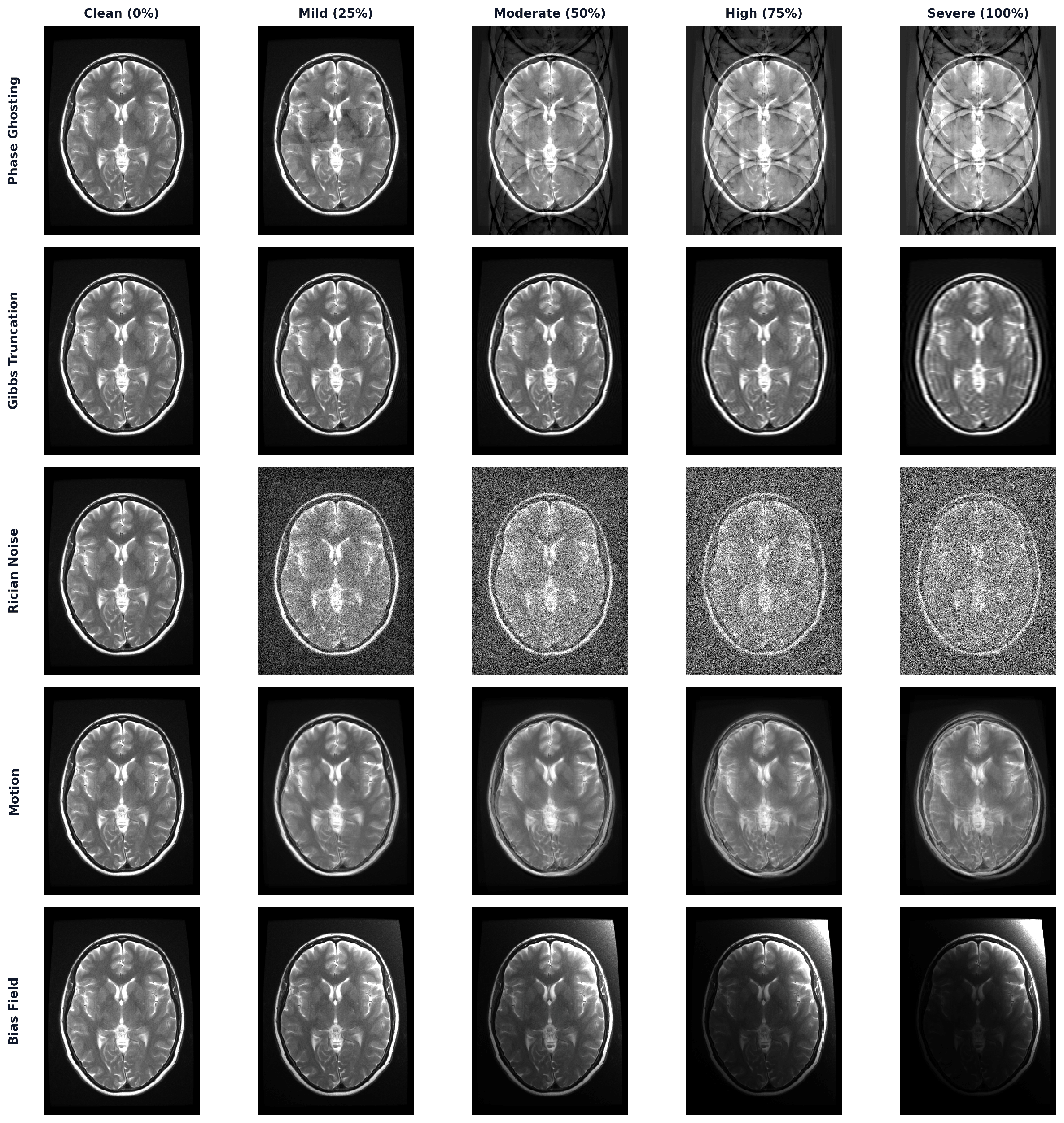}
\caption{\textbf{Multi-Artifact Physical Severity Examples.} Representative axial slices demonstrating the five simulated MRI artifact classes across five severity tiers: Clean ($0\%$), Mild ($25\%$), Moderate ($50\%$), High ($75\%$), and Severe ($100\%$). Row 1: Phase ghosting. Row 2: Gibbs truncation. Row 3: Rician noise. Row 4: Rigid motion. Row 5: $B_1$ receive bias field.}
\label{fig:dose-gallery}
\end{figure}

\subsection{Common exemplar metadata descriptions}
Supplementary Table \ref{tab:top_10_metadata_verbatim} presents the five most frequently occurring metadata descriptions in the clinical dataset. Each entry combines structured DICOM-derived information, including imaging plane, scanner configuration (manufacturer, model, and field strength), acquisition protocol details (description, sequence, and variant) and key imaging parameters such as echo time (TE), repetition time (TR), inversion time (TI) (in seconds), and flip angle (in degrees).

\begin{table}[ht]
\centering
\renewcommand{\arraystretch}{1.0}
\setlength{\tabcolsep}{5pt}
\caption{Top 5 most common metadata descriptions in the clinical dataset.}
\label{tab:top_10_metadata_verbatim}
\begin{tabular}{p{14.5cm}}
\midrule
A brain MRI, \textbf{\textit{plane}} axial, \textbf{\textit{Scanner}} (\textbf{\textit{Manufacturer}}, \textbf{\textit{Model}}, \textbf{\textit{Field Strength}}): (GE, Signa\_HDxt, 1.5), \textbf{\textit{Acquisition}} (\textbf{\textit{Description}}, \textbf{\textit{Sequence}}, \textbf{\textit{Variant}}): (Ax T2 FLAIR, SE\_IR, SK), \textbf{\textit{Imaging Parameters}} (\textbf{\textit{Echo Time}}, \textbf{\textit{Repetition Time}}, \textbf{\textit{Inversion Time}}, \textbf{\textit{Flip Angle}}): (0.129, 8.002, 2, 90) \\
\midrule
A brain MRI, \textbf{\textit{plane}} axial, \textbf{\textit{Scanner}} (\textbf{\textit{Manufacturer}}, \textbf{\textit{Model}}, \textbf{\textit{Field Strength}}): (GE, Signa\_HDxt, 1.5), \textbf{\textit{Acquisition}} (\textbf{\textit{Description}}, \textbf{\textit{Sequence}}, \textbf{\textit{Variant}}): (Ax PD/T2, SE, SK), \textbf{\textit{Imaging Parameters}} (\textbf{\textit{Echo Time}}, \textbf{\textit{Repetition Time}}, \textbf{\textit{Inversion Time}}, \textbf{\textit{Flip Angle}}): (0.0203, 5.9, NONE, 90) \\
\midrule
A brain MRI, \textbf{\textit{plane}} axial, \textbf{\textit{Scanner}} (\textbf{\textit{Manufacturer}}, \textbf{\textit{Model}}, \textbf{\textit{Field Strength}}): (Siemens, Aera, 1.5), \textbf{\textit{Acquisition}} (\textbf{\textit{Description}}, \textbf{\textit{Sequence}}, \textbf{\textit{Variant}}): (t2\_tse\_tra\_384\_p2, SE, SK\_SP\_OSP), \textbf{\textit{Imaging Parameters}} (\textbf{\textit{Echo Time}}, \textbf{\textit{Repetition Time}}, \textbf{\textit{Inversion Time}}, \textbf{\textit{Flip Angle}}): (0.087, 3.96, NONE, 150) \\
\midrule
A brain MRI, \textbf{\textit{plane}} axial, \textbf{\textit{Scanner}} (\textbf{\textit{Manufacturer}}, \textbf{\textit{Model}}, \textbf{\textit{Field Strength}}): (Siemens, Avanto, 1.5), \textbf{\textit{Acquisition}} (\textbf{\textit{Description}}, \textbf{\textit{Sequence}}, \textbf{\textit{Variant}}): (pd+t2\_tse\_tra, SE, SK\_SP\_OSP), \textbf{\textit{Imaging Parameters}} (\textbf{\textit{Echo Time}}, \textbf{\textit{Repetition Time}}, \textbf{\textit{Inversion Time}}, \textbf{\textit{Flip Angle}}): (0.014, 3.68, NONE, 150) \\
\midrule
A brain MRI, \textbf{\textit{plane}} coronal, \textbf{\textit{Scanner}} (\textbf{\textit{Manufacturer}}, \textbf{\textit{Model}}, \textbf{\textit{Field Strength}}): (GE, SIGNA\_HDx, 1.5), \textbf{\textit{Acquisition}} (\textbf{\textit{Description}}, \textbf{\textit{Sequence}}, \textbf{\textit{Variant}}): (Cor T1, SE, NONE), \textbf{\textit{Imaging Parameters}} (\textbf{\textit{Echo Time}}, \textbf{\textit{Repetition Time}}, \textbf{\textit{Inversion Time}}, \textbf{\textit{Flip Angle}}): (0.02, 0.4, NONE, 90) \\
\bottomrule
\end{tabular}
\end{table}

\end{appendices}

\newpage
\bibliography{sn-bibliography}

\end{document}